\documentclass[10pt,journal,compsoc]{IEEEtran}
\usepackage[table,dvipsnames]{xcolor}
\usepackage[table]{xcolor}
\RequirePackage[dvipsnames]{xcolor}
\usepackage{amsmath,amsfonts}
\usepackage{algorithmic}
\usepackage{array}
\usepackage[caption=false,font=normalsize,labelfont=sf,textfont=sf]{subfig}
\ifCLASSOPTIONcaptionsoff
 \usepackage[nomarkers]{endfloat}
\let\MYoriglatexcaption\caption
\renewcommand{\caption}[2][\relax]{\MYoriglatexcaption[#2]{#2}}
\fi
\usepackage{textcomp}
\usepackage{stfloats}
\usepackage{url}
\usepackage{verbatim}
\usepackage{graphicx}
\def\BibTeX{{\rm B\kern-.05em{\sc i\kern-.025em b}\kern-.08em
    T\kern-.1667em\lower.7ex\hbox{E}\kern-.125emX}}
\usepackage{balance}
\usepackage[caption=false,font=normalsize,labelfont=sf,textfont=sf]{subfig}
\usepackage{textcomp}
\usepackage{stfloats}
\usepackage{url}
\usepackage{verbatim}
\usepackage{graphicx}
\ifCLASSOPTIONcompsoc
  \usepackage[nocompress]{cite}
\else
  \usepackage{cite}
\fi
\usepackage{color}
\usepackage[normalem]{ulem}  
\usepackage{xspace}  
\usepackage{booktabs}
\usepackage{multirow}
\usepackage{pifont}
\usepackage{arydshln}
\usepackage{colortbl}
\usepackage{makecell,rotating,multirow,diagbox}
\usepackage{xspace}

\ExplSyntaxOn
\cs_new_protected:Npn \onedot {%
  \peek_meaning:NTF . 
    { }               
    { . }             
  \xspace             
}
\ExplSyntaxOff

\def\ie{\emph{i.e}\onedot}

\begin{document}

\title{Implicit Geometry Representations for Vision-and-Language Navigation from Web Videos}


\author{Mingfei~Han*\thanks{*Authors contribute equally.},
        Haihong~Hao*,
        Liang Ma,
        Kamila Zhumakhanova, 
        Ekaterina Radionova, 
        Jingyi Zhang, 
        Xiaojun Chang, 
        Xiaodan Liang, 
        Ivan Laptev
\IEEEcompsocitemizethanks{
\IEEEcompsocthanksitem Mingfei Han, Liang Ma, Kamila Zhumakhanova, Ekaterina Radionova, Xiaojun Chang, Xiaodan Liang and Ivan Laptev are with the Department of Computer Vision, Mohamed Bin Zayed University of Artificial Intelligence, Abu Dhabi, United Arab Emirates. \protect \\
E-mail: \{mingfei.han, liang.ma\}@mbzuai.ac.ae
\IEEEcompsocthanksitem Haihong Hao is with School of Information Science and Technology, University of Science and Technology of China. \protect
\IEEEcompsocthanksitem Jingyi Zhang is with Shenzhen Campus of Sun Yat-Sen University. \protect 

}}

\markboth{Submission to IEEE Transactions on Pattern Analysis and Machine Intelligence}%
{Shell \MakeLowercase{\textit{et al.}}: A Sample Article Using IEEEtran.cls for IEEE Journals}


\IEEEtitleabstractindextext{%
   \begin{abstract}
Vision-and-Language Navigation (VLN) has long been constrained by the limited diversity and scalability of simulator-curated datasets, which fail to capture the complexity of real-world environments. To overcome this limitation, we introduce a large-scale video-instruction framework derived from web-based room tour videos, enabling agents to learn from natural human walking demonstrations in diverse, realistic indoor settings. Unlike existing datasets, our framework integrates both \textbf{open-ended description-enriched trajectories} and \textbf{action-enriched trajectories} reconstructed in 3D, providing richer spatial and semantic supervision. A key extension in this work is the incorporation of \textbf{implicit geometry representations}, which extract spatial cues directly from RGB frames without requiring fragile 3D reconstruction. This approach substantially improves data utilization, alleviates reconstruction failures, and unlocks large portions of previously unusable video data. Comprehensive experiments across multiple VLN benchmarks (CVDN, SOON, R2R, and REVERIE) demonstrate that our method not only sets new state-of-the-art performance but also enables the development of robust zero-shot navigation agents. By bridging large-scale web videos with implicit spatial reasoning, this work advances embodied navigation towards more scalable, generalizable, and real-world applicable solutions.

\end{abstract}

}

\maketitle

\IEEEdisplaynontitleabstractindextext

%
\IEEEpeerreviewmaketitle
 
\section{Introduction}

\IEEEPARstart{O}{ver} the past few years, the field of Vision-and-Language Navigation (VLN)~\cite{reverie,r2r,wang2020environment,jain2019stay,liang2024cornav,ku2020room} has primarily relied on manually designed simulators and annotated navigation paths. Early benchmarks such as R2R~\cite{r2r} pioneered language-guided navigation within simulated indoor environments, while later datasets—including CVDN~\cite{cvdn}, REVERIE~\cite{reverie}, and SOON~\cite{soon}—extended VLN to dialogue-based and object-centric tasks. Despite these advances, the heavy dependence on simulated environments limits the diversity of visual scenes and fails to capture the complexity, clutter, and variability of real-world settings.

To mitigate the lack of diversity in training data, several recent efforts have attempted to construct richer, more complex datasets. AirBERT~\cite{airbert} leverages Airbnb imagery to synthesize panoramic views, yet these static images lack the temporal coherence and spatial continuity needed for navigation. ScaleVLN~\cite{scalevln} utilizes curated 3D environments~\cite{xia2018gibson,ramakrishnan2021habitat} for large-scale training, but faces high reconstruction costs and scalability limitations. YTB-VLN~\cite{youtube_vln} extracts panoramic views and text from web videos, while NaVid~\cite{zhang2024navid} integrates MatterPort3D~\cite{Matterport3D} with video data for sim-to-real transfer. Nevertheless, these datasets still fall short of jointly achieving (1) \textbf{broad scene diversity}, (2) \textbf{open-vocabulary object variety}, and (3) \textbf{comprehensive spatial understanding}—three essential ingredients for developing robust, generalizable embodied agents.

To bridge this gap, we introduce \textbf{RoomTour3D}, a large-scale dataset that provides spatially enriched, geometry-aware supervision for VLN. RoomTour3D leverages publicly available room tour videos, which naturally capture human walking trajectories through diverse real-world indoor environments. These first-person-view videos offer agent-centric perspectives that encompass a broad spectrum of room layouts, furniture arrangements, and lighting conditions. The continuous motion within each video provides dense spatial context and multi-view coverage, making them ideal for modeling realistic navigation behaviors.

To fully harness the potential of such videos, we develop an automated pipeline that extracts open-ended, geometry-aware human walking trajectories and generates spatially grounded navigation instructions in open-vocabulary form. Our pipeline densely samples frames along continuous walk-through sequences to model realistic navigation scenarios. Using COLMAP~\cite{schoenberger2016sfm,schoenberger2016mvs}, we reconstruct 3D environments to recover geometric cues such as camera poses and orientations. Decision-making frames are sampled at turning points with maximal yaw rotation, while additional frames are selected approximately every 1.5 meters to define navigable steps. The reconstructed scenes are further enriched using RAM~\cite{zhang2023recognize} for object recognition, Grounding-DINO~\cite{liu2023grounding} for object localization, and Depth-Anything~\cite{depthanything} for depth estimation. Finally, GPT-4~\cite{openai2022gpt4} generates open-vocabulary navigation instructions, enabling both descriptive summarization and task-specific guidance.

While explicit 3D reconstruction provides valuable spatial structure, it remains fragile and inefficient in real-world web videos. In our preliminary data collection, nearly 200K candidate trajectories were gathered, but only about 17K could be successfully reconstructed—resulting in over 90\% data loss. This low success rate arises from inherent challenges in web footage, such as motion blur, dynamic objects, inconsistent lighting, and insufficient scene overlap. Consequently, explicit geometric reconstruction severely limits scalability and data utilization.

To address this limitation, we introduce \textbf{implicit geometry representations (IGR)} as a key extension of RoomTour3D. Instead of relying on explicit 3D reconstruction, IGR extracts spatial cues directly from RGB frames using learned spatial encodings. This implicit approach preserves essential geometric information without the need for fragile structure-from-motion pipelines. Inspired by advances in implicit geometry learning such as VLM-3R~\cite{fan2025vlm} and Cut3R~\cite{wang2025continuous}, our design effectively leverages previously discarded video data, substantially expanding usable training samples. Within VLN tasks, implicit geometry encodes rich spatial priors that enhance decision-making and path planning while remaining robust to visual noise and environmental variability.

Our resulting dataset comprises approximately \textbf{100K open-ended trajectories}, \textbf{200K descriptive captions}, and \textbf{17K action-enriched, geometry-aware trajectories} collected from \textbf{1,847 homes}. We additionally release intermediate annotations—object tags, bounding boxes, depth maps, room categories—and the corresponding code and prompts for instruction generation. Extensive experiments with NaviLLM~\cite{navillm}, a state-of-the-art language model–based navigation agent, demonstrate consistent performance gains exceeding 6\% across multiple VLN benchmarks (CVDN, SOON, R2R, and REVERIE), including a 9.8\% improvement on SOON. When enhanced with implicit geometry, RoomTour3D-IGR achieves an additional 8\% gain, confirming its ability to improve navigation robustness and scalability.

In summary, this paper makes the following contributions:
\begin{itemize}
\item We curate a novel dataset derived from diverse room tour videos, which provides longer trajectories, more complex environments, and greater scene continuity than previous web-based datasets.
\item We introduce a pipeline to extract geometry-aware navigation instructions, aligning spatial understanding with navigation goals, and generate open-vocabulary instructions for flexible and real-world navigation tasks.
\item We propose the use of implicit spatial encodings, which replace traditional explicit geometry, offering significant improvements in data utilization and robustness.
\item We conduct thorough experiments and ablation studies, demonstrating how our dataset and method improve the performance of state-of-the-art models, enabling zero-shot generalization across diverse VLN tasks.
\end{itemize}

Building upon our prior work~\cite{han2025roomtour3d}, this extension incorporates RoomTour3D-IGR—a version of RoomTour3D utilizing implicit geometry representations. This results in substantial improvements in three major areas:
\begin{itemize}
\item We replace explicit geometric reconstruction with implicit spatial encodings, overcoming issues associated with noisy or incomplete geometry data from web-based videos, reducing the need for data filtering and reconstruction failures.
\item Our extensive ablation studies demonstrate that implicit geometry effectively reutilizes a significant portion of previously discarded data, thus broadening the available training set.
\item Comprehensive experiments across multiple VLN tasks show that our approach boosts performance while broadening RoomTour3D's applicability and lowering barriers to data collection, making it a practical solution for enhancing navigation agents using web-based videos.
\end{itemize}

\section{Related Work}

\subsection{Vision-and-Language Navigation}

Navigating unfamiliar indoor environments through natural language commands is a core capability for embodied agents designed to assist humans. The overarching goal of Vision-and-Language Navigation (VLN) is to enable agents to interpret linguistic instructions and complete diverse navigation tasks~\cite{r2r,soon}. Over time, a variety of benchmarks have been introduced to capture different aspects of this challenge. The R2R dataset~\cite{r2r} pioneered fine-grained, step-by-step instruction following in simulated environments. CVDN~\cite{cvdn} expanded the paradigm to dialogue-driven navigation, while REVERIE~\cite{reverie} and SOON~\cite{soon} focused on grounding natural language instructions to object-centric goals. Related lines of work such as Embodied Question Answering~\cite{eqa_1,eqa_2} further extended these tasks toward active exploration in 3D environments.

Recent advances have explored multiple strategies for improving vision–language grounding and action reasoning. NavGPT2~\cite{zhou2025navgpt} employs a Q-former for fine-grained cross-modal alignment, augmented with topological maps to support spatial reasoning. NavCoT~\cite{lin2024navcot} adapts LLaMA-7B with Chain-of-Thought reasoning to enable text-only large language models to perform embodied navigation. SAME~\cite{zhou2024same} introduces a State-Adaptive Mixture-of-Experts framework to handle dynamic observations, while SRDF~\cite{wang2024bootstrapping} scales instruction–trajectory pairs through automatic generation, achieving human-level or better results. Other studies explore causal modeling (GOAT~\cite{wang2024vision}), depth-augmented exploration (SUSA~\cite{zhang2024agent}), and BEV-style reasoning (BSG~\cite{liu2023bird}) inspired by multi-camera perception in autonomous driving~\cite{li2022bevformer}.

While these methods collectively advance VLN, most still depend heavily on 2D visual cues and are often optimized for narrow, task-specific benchmarks~\cite{hamt,duet,gao2023adaptive,kerm,hwang2023meta,vln_bert,long2023discuss,liu2024volumetric,Wang2024GOAT,zhao2024overnavelevatingiterativevisionandlanguage,metaexplore,soon,hao2020towards,qiao2022hop,li2023improving,qiao2023vln,an2023bevbert,hong20233dllm}. This specialization often limits their generalization to unseen environments~\cite{hao2025conav}. In contrast, NaviLLM~\cite{navillm} demonstrates a more generalist framework, training a single LLM-based model jointly across multiple VLN tasks and achieving strong cross-task transferability. Nevertheless, even such generalist approaches remain constrained by dataset limitations—most existing benchmarks lack real-world diversity and rely on explicit 3D annotations that are both costly and fragile. Our RoomTour3D dataset addresses these bottlenecks by leveraging large-scale, web-sourced room tour videos that provide continuous, richly annotated trajectories in realistic indoor environments.

\subsection{Data-Centric Methods for VLN}

A persistent challenge in VLN research is the \textbf{limited availability of large, diverse training data}, which restricts agents’ ability to generalize to unseen environments. Widely used datasets such as R2R~\cite{r2r}, RxR~\cite{ku2020room}, CVDN~\cite{cvdn}, and SOON~\cite{soon} are all simulator-based and thus expensive to expand due to their heavy reliance on manual annotation. To mitigate this, several studies have explored data augmentation~\cite{aug1,aug2,aug3,aug4,aug5,aug6,aug7} and self-exploration in virtual environments~\cite{self_explore1,self_explore2}, which improve sample efficiency but do not fundamentally increase data diversity.

Efforts to move beyond handcrafted simulators include web-based and synthetic approaches. VLN-BERT~\cite{vln_bert} and AirBERT~\cite{airbert} pretrain models on web image–caption pairs to improve vision–language alignment; however, such image collections lack the spatial and temporal continuity necessary for realistic navigation. Automated data generation pipelines~\cite{hm3d_auto,new_path}, such as ScaleVLN~\cite{scalevln}, leverage curated 3D scenes or synthetic environments but remain limited by reconstruction cost and lack of photorealistic variability. PanoGen~\cite{li2023panogen} expands data diversity by synthesizing panoramic indoor scenes using diffusion models and recursive outpainting, though synthetic artifacts reduce realism. YTB-VLN~\cite{youtube_vln} scales data collection by using YouTube room tour videos, yet its trajectories are discretized into panoramic nodes and lack explicit geometric reasoning, hindering applicability to embodied agents and real-world robotics.

Our RoomTour3D dataset overcomes these constraints by departing from template-based pipelines. It introduces: (i) open-vocabulary, free-form instructions generated by large language models, (ii) open-ended trajectories extracted directly from continuous human walking videos, and (iii) decision-point and spatially proximate frame sampling for defining navigable actions. Combined with 3D reconstruction and language-based scene descriptions, RoomTour3D offers a richer and more realistic supervision signal, bringing VLN training closer to real-world navigation behavior.

\subsection{Geometric Information from Monocular Images}

Inferring 3D structure from monocular imagery is a long-standing challenge in computer vision~\cite{wang2025continuous}. Existing methods can be broadly categorized into geometry-based and learning-based approaches.

Classical geometry-based methods reconstruct 3D scenes explicitly using only observed image sequences~\cite{wang2025continuous}. Structure-from-Motion (SfM)~\cite{agarwal2010bundle,agarwal2011rome,hartley2003multiple,lowe2004distinctive,schonberger2016sfm,seitz2006comparison,snavely2006photo,snavely2008modeling,triggs2000bundle} estimates camera poses and sparse scene geometry by matching features across views, while Simultaneous Localization and Mapping (SLAM)~\cite{cadena2016past,davison2007monoslam,durrant2006slam1,engel2014lsdslam,klein2007ptam,mur2015orbslam,newcombe2011dtam} incrementally builds spatial maps while tracking camera motion. More recently, neural rendering methods—such as NeRF~\cite{mildenhall2021nerf,chen2022tensorf,fridovich2022plenoxels,wang2021neus,mueller2022instant} and Gaussian Splatting~\cite{kerbl2023gaussiansplatting}—have enabled photorealistic 3D reconstruction and novel-view synthesis. However, these pipelines rely on stable textures, consistent lighting, and dense viewpoint overlap—conditions rarely met in unconstrained web videos, where motion blur, dynamic objects, and abrupt camera movements are common.

Learning-based approaches incorporate prior knowledge to achieve more robust geometric inference, often through implicit geometry representations that capture spatial information directly from image embeddings. DUSt3R~\cite{wang2024dust3r} predicts dense point maps from uncalibrated image pairs, while CUT3R~\cite{wang2025continuous} extends this to streaming inputs for continuous online 3D estimation. MoGe~\cite{wang2025moge} reconstructs geometry from a single image via affine-invariant point maps. Recently, VGGT~\cite{wang2025vggt} employs a transformer architecture to jointly predict multiple 3D attributes—depth, pose, and correspondence—without iterative optimization, significantly improving efficiency and robustness. Unlike classical methods, VGGT effectively handles low-texture regions and large viewpoint changes, making it particularly suitable for web videos and large-scale, in-the-wild data.

These advancements directly motivate our use of implicit geometry representations in RoomTour3D. By replacing brittle explicit reconstruction with learned spatial embeddings, we can fully exploit large volumes of web video data that were previously unusable due to reconstruction failures. This shift not only enhances robustness to noise and incomplete geometry but also scales spatial reasoning to diverse, unconstrained real-world environments.

\begin{figure*}[ht!]
\includegraphics[width=0.98\linewidth]{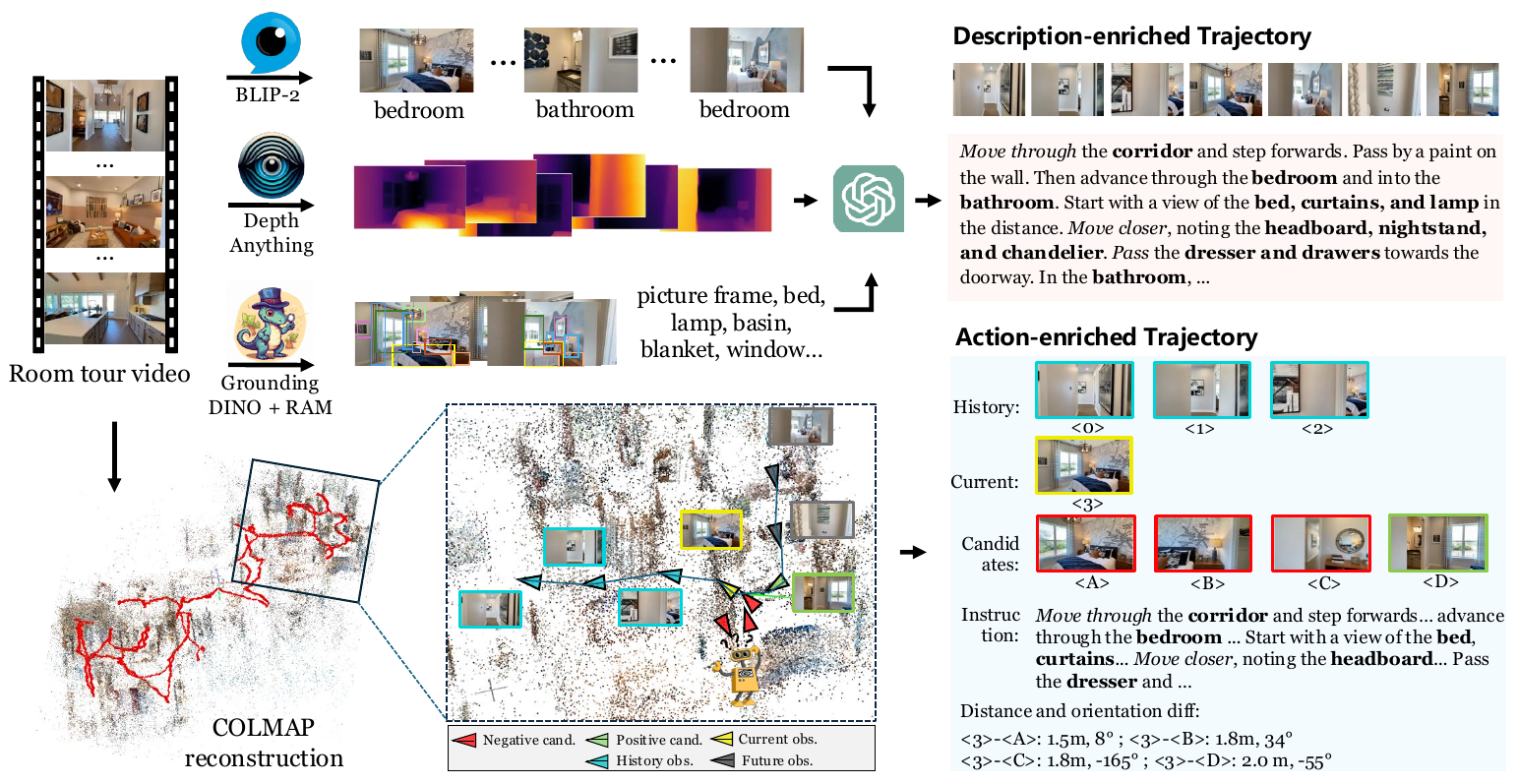}
\caption{\small Overview of our origin RoomTour3D with COLMAP reconstruction and explicit geometry information. 
Starting from a room tour video, we first apply BLIP-2~\cite{li2023blip2} on frame sequence to predict the room locations. Next, we use RAM~\cite{zhang2023recognize} and Grounding-DINO~\cite{liu2023grounding} to identify objects within the frames and employ Depth-Anything~\cite{depthanything} for depth prediction. Subsequently, COLMAP is used to reconstruct the 3D scene with complete geometry information, and we sample human walking trajectories from the continuous frames.
The trajectory captures open-world objects, their positions, and depths relative to the camera. Finally, we use advanced LLM, \ie, GPT-4 to generate the free-form descriptions for pretraining, namely description-enriched trajectories. Specifically, 
for the trajectory shown in the figure,
which involves instant turning points, we specially treat \textless 0\textgreater\ to \textless 6\textgreater\ as walking trajectory, \textless A\textgreater\, \textless B\textgreater\, and \textless C\textgreater\ as side-watching points and use them as negative candidates for navigation finetuning task, namely action-enriched trajectories. For more details, please refer to Section~\ref{sec:data_generatopm}.}
\label{fig:overall_pipeline}
\end{figure*}

\section{RoomTour3D} 
\label{sec:data_generatopm}

In this section, we describe the automatic data curation pipeline of RoomTour3D. The pipeline encompasses the complete annotation process—from sampling open-ended human walking trajectories to generating corresponding textual descriptions enriched with open-world object diversity and spatial awareness. Leveraging reconstructed 3D environments, we further sample navigable trajectories with associated actions. An overview of the entire data-generation process is presented in Figure~\ref{fig:overall_pipeline}.

\subsection{Room Tour Video Collection}
\label{sup:vid_collection}

To achieve high diversity in indoor environments, we exploit the abundance and variety of room tour videos available on YouTube. These videos, typically recorded with hand-held cameras from a first-person perspective, provide realistic, dynamic visual streams that closely resemble embodied viewpoints. Our final dataset comprises \textbf{1,847 room tour videos} totaling \textbf{243 hours} of footage. The collection process builds upon the list of videos used in YTB-VLN~\cite{youtube_vln}, which we subsequently refined and expanded to ensure greater scene variety and data quality.

To guarantee the usability of the videos for 3D reconstruction, we prioritized continuous recordings with minimal interruptions—avoiding human interviews, abrupt cuts, or close-up shots. We applied a title–description–based filtering strategy using GPT-4~\cite{openai2022gpt4} and excluded clips shorter than three minutes. Furthermore, we implemented a transition detector to identify abrupt scene changes and retained only those with at least nine continuous shots covering over 80\% of the video’s duration. To maintain dataset freshness and diversity, we continually monitor and update high-quality YouTube channels (e.g., NavaRealtyGroup, Open House 24, and Sona Visual) to include new uploads.

For processing efficiency, all videos are spatially downscaled to a shorter side of 360 pixels and temporally downsampled to three frames per second. All subsequent data processing is performed on these normalized sequences.

\subsection{Navigable Points and Instruction Generation}
\label{sup:nav_points}

\textbf{Navigable Points Generation.} To infuse open-world knowledge from room tour videos into navigation agents, we interpret individual video frames as potential navigable observations. Each frame in a human walking trajectory can be regarded as having two possible actions: move forward or stop. At points of significant view change—instances where the view direction shifts sharply within a small spatial radius—we extract frames with distinct orientations as candidate navigable actions. Unlike YTB-VLN~\cite{youtube_vln}, which constructs panoramic images at discrete room nodes, our approach directly samples significant view-change points and their neighboring frames based on geometric cues.

We detect such points by reconstructing the 3D scene to compute camera orientation differences and spatial distances between consecutive frames. Situations in which a person revisits nearly the same spatial position from a different viewpoint, or turns sharply within close proximity, are particularly informative for generating diverse navigation actions.

To identify these points, we calculate pairwise cosine similarity along the trajectory and retain frames exceeding a 45 degrees angular change. Non-maximum suppression is then applied to highlight local maxima in orientation shifts, isolating the most distinct transitions. Subsequently, DBSCAN clustering~\cite{dbscan} groups nearby but visually distinct frames, ensuring a set of diverse, spatially meaningful navigable actions even without panoramic synthesis.

\textbf{Instruction Generation.} We next transform the collected visual–spatial information into structured textual inputs suitable for GPT-based instruction generation. This process aggregates multi-source signals from RAM (Swin-L)~\cite{zhang2023recognize}, Grounding DINO~\cite{liu2023grounding}, and Depth-Anything~\cite{depthanything}, which collectively provide object categories, spatial coordinates, and depth cues.

\noindent\textbf{Object Variety to Text.} Room tour videos capture open-world indoor diversity—spanning furniture, décor, and layout variations—offering valuable context for training navigation models. We use RAM~\cite{zhang2023recognize} to extract object tags in each frame and remove entries corresponding to room types to maintain consistency with BLIP-2’s predicted room labels. These object tags are later used for spatial grounding and scene description.

\noindent\textbf{Spatial Awareness to Text.} Since navigation agents must reason about object positions and movements, spatial cues are vital. We jointly employ Grounding DINO~\cite{liu2023grounding} and Depth-Anything~\cite{depthanything} to estimate spatial localization and relative distance. Depth-Anything is preferred over COLMAP~\cite{schoenberger2016sfm,schoenberger2016mvs} depth because it directly provides dense and reliable depth maps without dependence on structure-from-motion, which is often error-prone in unconstrained web videos.

Each frame is divided horizontally into three spatial zones—left (30\%), center (40\%), and right (30\%)—and depth is quantized into near (30\%), medium (40\%), and far (30\%) ranges. Overlap ratios between object bounding boxes and these regions determine relative spatial positions. For example, an object overlapping more than 30\% with the near zone is labeled in the near distance. Large objects spanning multiple ranges (e.g., carpets) are annotated across all relevant zones. This structured description encodes both position and scale, forming a rich textual input for GPT-based summarization.

\noindent \textbf{GPT-Based Instruction Synthesis.} We employ GPT-4~\cite{openai2022gpt4} to summarize the evolving object configurations along each walking trajectory. Object tags, spatial relations, and room locations are organized per frame and then formatted into a unified prompt template. GPT-4 synthesizes these structured inputs into coherent, context-aware navigation instructions, yielding high-quality training text for VLN agents.

\subsection{Enriched Trajectories}
\begin{figure*}[ht]
\includegraphics[width=\linewidth]{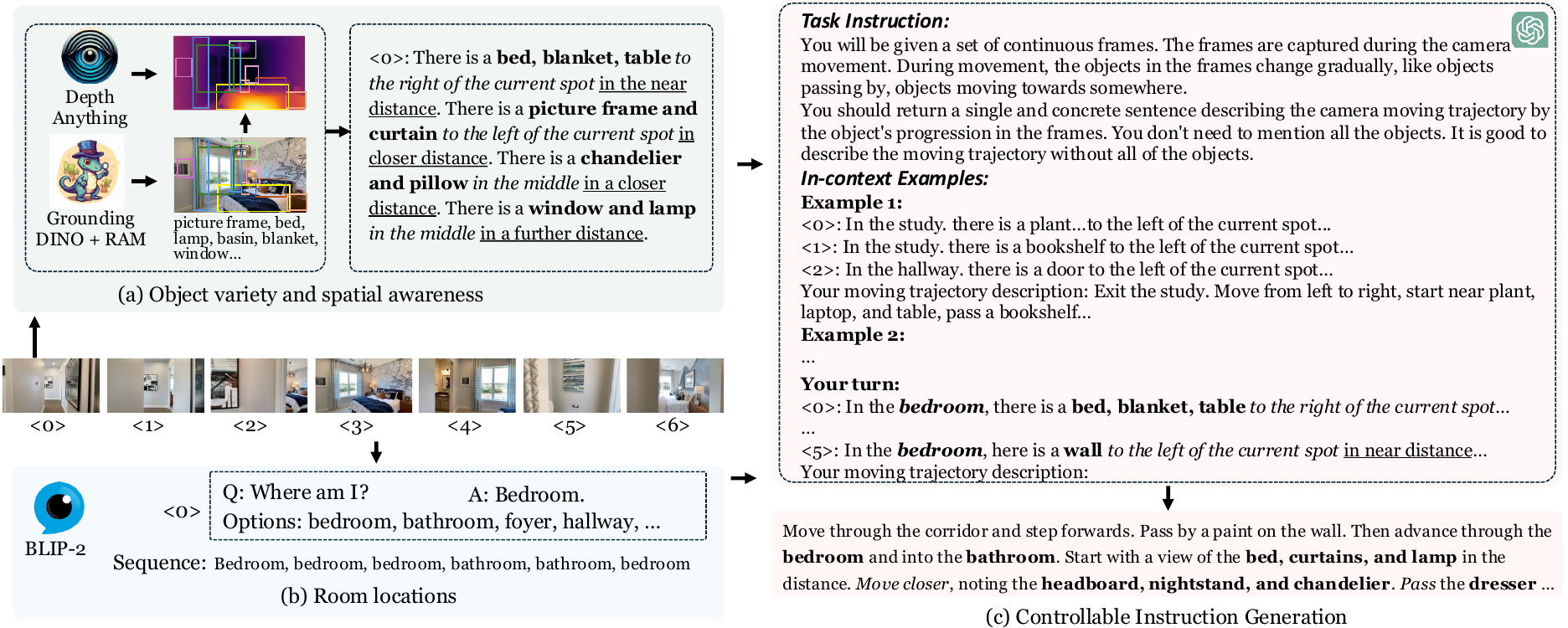}
\caption{Instruction generation in a controllable way. (a) Using open-sourced expert models, we identify \textit{what objects are in the frames}, and assess \textit{how far an object is} and determine \textit{where an object is located}. The information is then textualized to create richly detailed frame captions. (b) BLIP-2 is adopted to predict and smooth room location across sequential frames. (c) Combining room locations and object information, we use GPT-4 for controllable and open-vocabulary instruction generation. The prompt consists of a task instruction that defines the generation task, and in-context examples that constrain the output style.}
\label{fig:generation_detail}
\end{figure*}

\subsubsection{Description-Enriched Trajectories}
This subsection describes the generation of controllable, text-based descriptions for open-ended trajectories. We begin by uniformly sampling human-walking sequences at a rate of one frame every two seconds—approximately corresponding to an average walking speed of 1.42~m/s~\cite{wikiwalking}, which is typically slower indoors. As illustrated in Figure~\ref{fig:generation_detail}, each trajectory is annotated using expert models including BLIP-2~\cite{li2023blip2}, RAM~\cite{zhang2023recognize}, Grounding-DINO~\cite{liu2023grounding}, and Depth-Anything~\cite{depthanything}, which collectively provide information about object categories, spatial locations, and depth. The extracted multimodal cues are then composed into structured text and integrated with GPT-4~\cite{openai2022gpt4} to generate coherent and detailed trajectory-level instructions.

\noindent\textbf{Object Variety and Spatial Awareness.}
To encode both semantic and spatial richness, we combine outputs from three expert models within a textual template: ``There is a \textit{[object tag]} to the \textit{[spatial position]} of the current spot in \textit{[relative distance]}.’’ Specifically, RAM~\cite{zhang2023recognize} identifies object categories, Grounding-DINO~\cite{liu2023grounding} localizes objects spatially, and Depth-Anything~\cite{depthanything} estimates relative depth.  
By analyzing object bounding-box centers and corresponding depth regions, we determine spatial placement with respect to the camera. The frame is divided into three lateral zones—left (30\%), center (40\%), and right (30\%)—and depth is categorized as near (closest 30\%), medium (next 40\%), or far (remaining 30\%). Overlap ratios between bounding boxes and these defined zones establish spatial and depth annotations. For instance, an object overlapping more than 30\% with the near region is labeled as being in the near distance. Larger items spanning multiple zones, such as carpets or tables, are annotated across all relevant categories to preserve their extended spatial footprint.  
This procedure, visualized in Figure~\ref{fig:generation_detail}(a), ensures the resulting captions accurately encode both object variety and spatial relationships, facilitating effective instruction generation.

\noindent\textbf{Room Location Annotation in Videos.}
To determine frame-level room categories, we employ BLIP-2~\cite{li2023blip2} in visual question–answering mode, using the prompt ``Which room am I in?’’ Possible responses are restricted to a predefined list of 16 common room types (\textit{bathroom, kitchen, bedroom, living room, hallway, office, dining room, foyer, laundry room, outside, porch, garage, front, patio, driveway, and backyard}). This taxonomy is constructed by analyzing ten randomly sampled long videos, where BLIP-2 in generative mode identifies and ranks frequently mentioned room types. For per-frame labeling, we switch to BLIP-2’s discriminative mode and apply temporal smoothing to mitigate noise.  
As shown in Figure~\ref{fig:generation_detail}(b), this produces temporally consistent predictions. Manual evaluation on 50 clips achieved an accuracy of 85\%, validating the reliability of this method. Restricting the answer set simplifies classification and reduces ambiguity, while the subsequent GPT-based trajectory summarization restores open-vocabulary expressiveness.

\noindent\textbf{Controllable Instruction Generation.}
Finally, we integrate the room annotations with frame-level object descriptions and feed them into GPT-4-Turbo~\cite{openai2022gpt4} for open-vocabulary instruction generation. Following the ``Task Instruction – In-context Examples – Prediction’’ schema (see Figure~\ref{fig:generation_detail}(c)), GPT is guided to output concise and structured navigation-style descriptions. The in-context examples constrain style and tone, while the task prompt defines the goal—summarizing object progression and spatial transitions throughout the trajectory. The generated text provides coherent, controllable supervision suitable for training large-scale VLN models.

\subsubsection{Action-Enriched Trajectories}

\begin{figure*}[t!]
\includegraphics[width=\linewidth]{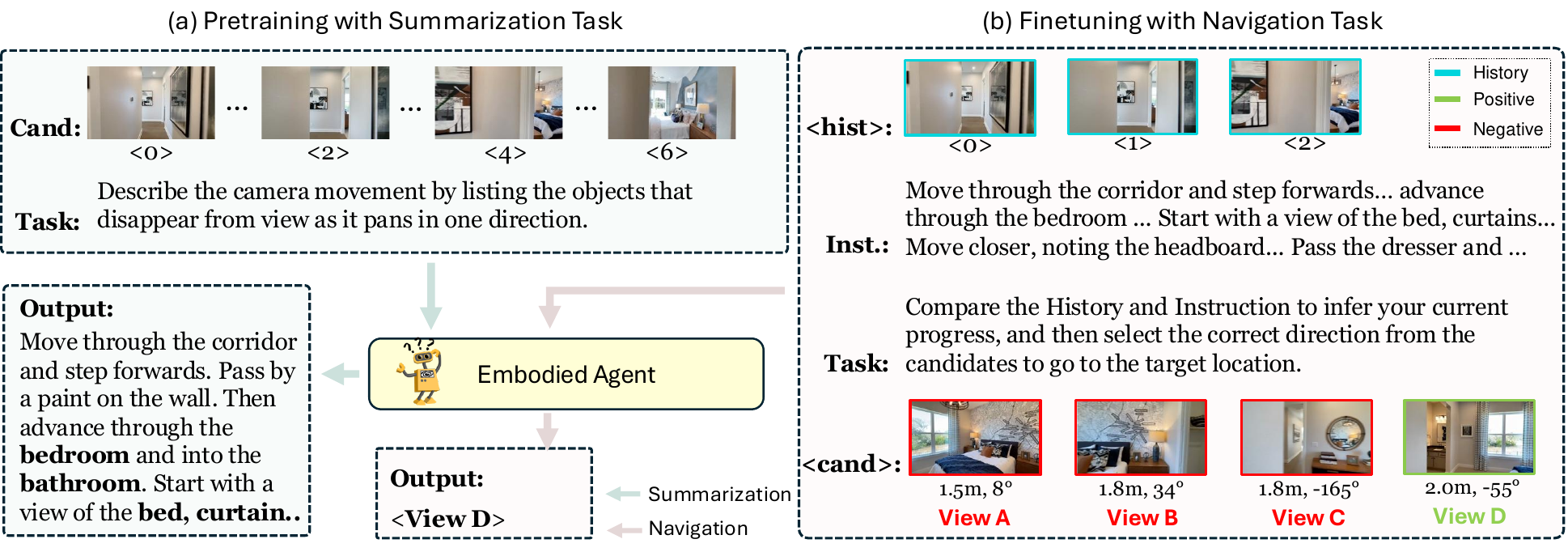}
\caption{Model training diagram using RoomTour3D. Two complementary tasks are designed to enhance NaviLLM: (a) \textbf{Pretraining.} Sampled frames along the trajectory serve as candidate observations, and the model is optimized to summarize object progression along the path. (b) \textbf{Finetuning.} Each frame acts as a navigable step. Given the historical observations \textless 0\textgreater–\textless 2\textgreater\ and a navigation instruction, the model predicts the next action by choosing among candidate views (A–D).}
\label{fig:method}
\end{figure*}

\noindent\textbf{3D Environment Reconstruction.}
To obtain accurate geometric information for each trajectory, we reconstruct 3D environments using COLMAP~\cite{schoenberger2016sfm,schoenberger2016mvs}. Videos are sampled at 3~frames/s to balance accuracy and efficiency, and are segmented into 100-second overlapping clips (with 10-second overlaps) to enable parallel reconstruction. The resulting sub-models are merged by aligning overlapping frames—two adjacent models are fused when at least three frames overlap.  
Because reconstruction quality may vary, multiple models may be generated from a single clip. To merge them systematically, we build a graph where each sub-model is represented as a node and edges indicate overlapping frames. A depth-first search~\cite{depth_search} is then applied to merge connected components recursively until a unified model is obtained.

\noindent\textbf{Navigable Action Sampling.}
Navigation actions are extracted by identifying significant view-change points within small spatial neighborhoods. The reconstructed 3D scenes allow estimation of camera orientation differences and spatial distances between frames, enabling detection of revisits and sharp turns. Frames exhibiting angular changes above 45° are retained via cosine similarity filtering. Non-maximum suppression is then applied to isolate the strongest local changes, and DBSCAN clustering~\cite{dbscan} groups spatially close yet visually distinct points.  
This procedure—visualized conceptually in Figure~\ref{fig:method}(b)—is robust to occasional reconstruction errors. Within each cluster, the most recent frame is designated as the positive candidate, and the frame with the greatest angular deviation serves as the negative candidate, producing diverse and realistic navigable actions.

\subsubsection{Data Correctness Verification}
To evaluate the reliability of the automatic data generation pipeline, we manually inspected 100 randomly selected trajectory descriptions. Each was scored on a four-point relevance scale (1 = \emph{totally irrelevant}, 2 = \emph{partially relevant}, 3 = \emph{mostly relevant}, 4 = \emph{perfect match}). The average rating was 3.08, with 74\% of samples judged ``mostly relevant’’ or ``perfect match,’’ demonstrating that the generated annotations are both accurate and visually consistent. Representative examples of this verification are provided in the appendix.

\subsection{From Explicit to Implicit}
Implicit geometry representations (IGR) have been extensively explored in recent learning-based 3D methods such as CUT3R~\cite{wang2025continuous} and VGGT~\cite{wang2025vggt}. Prior studies suggest that implicit energy fields naturally encode geometric smoothness, continuity, and global consistency without requiring dense discretization into meshes or point clouds. By shaping a continuous energy landscape instead of directly regressing explicit structures, such models effectively represent multimodal or uncertain geometries while maintaining overall coherence.  

Similar advantages are reported in robot manipulation, where energy-based implicit behavioral policies outperform explicit imitation strategies such as mean-squared error or mixture density networks~\cite{Tosi2021SMDNetsSM}. Their superiority arises from representing continuous energy landscapes across the action space, allowing optimization toward low-energy, high-quality decisions.  

Our preliminary experiments also support these observations: implicit geometry formulations capture global consistency even under irregular spatial sampling, enabling accurate 3D estimation without explicit continuity (see Figure~\ref{fig:ext1}). This property aligns with the discrete and unconstrained nature of web-based videos, which lack the temporal and spatial regularity of simulator environments. Motivated by these findings, we replace explicit geometric inputs in our VLN model with implicit geometry encodings. Although the output policy remains discrete, using implicit spatial inputs provides smoother, more coherent, and noise-resilient perception—benefits verified in our subsequent experiments.

\subsection{End-to-End RoomTour3D-IGR}
\begin{figure*}[t]
  \centering
  \includegraphics[width=0.96\linewidth]{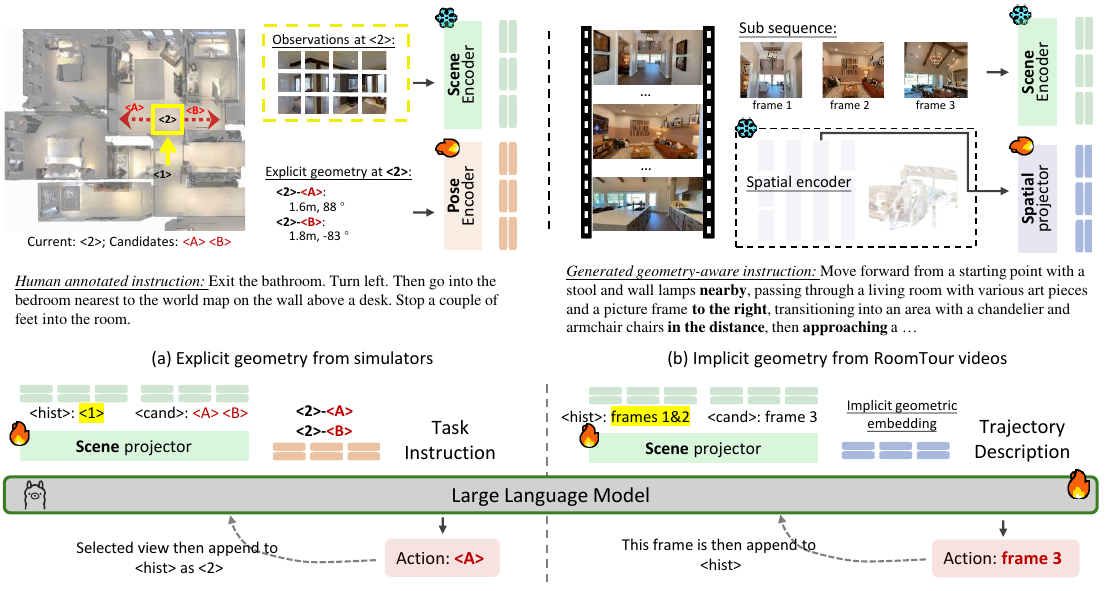}
  \vskip -0.2in
  \caption{Overview of implicit-geometry training. Our RoomTour3D-IGR processes both explicit geometry from simulators and implicit geometry from RoomTour videos, alongside task instructions. (a) For simulator data, the pipeline follows the original RoomTour3D setup: RGB observations are encoded by the scene encoder, while explicit geometric features (e.g., distance and heading) are incorporated. For instance, \textless2\textgreater–\textless A\textgreater: 1.6~m, 88° denotes a spatial relation guiding the agent’s next action. (b) For RoomTour videos, frames 1–2 form the trajectory history, and frame~3 acts as a navigation candidate. RGB frames are encoded by the scene encoder, while a VGGT-based spatial encoder extracts implicit geometric features. These embeddings are projected via a spatial projector into the LLM’s latent space to guide accurate action prediction.}
  \label{fig:ext1}
\end{figure*}

We introduce \textbf{RoomTour3D-IGR}, an extension of our navigation framework that employs implicit geometry representations. As illustrated in Figure~\ref{fig:ext1}, the architecture extends a pretrained navigation agent by integrating a new spatial encoding module capable of inferring geometry directly from RGB inputs. This design overcomes the principal limitation of web-based data—namely, the absence of explicit geometric annotations—allowing effective use of previously discarded videos.

\noindent\textbf{Spatial Encoder.}
Instead of relying on explicit pose-based geometric information, we adopt implicit representations extracted via a \emph{Spatial Encoder}. This encoder reuses the transformer blocks of the pretrained VGGT~\cite{wang2025vggt} model (excluding its prediction heads). Given a sequence of $N$ frames $(I_i)_{i=1}^N$, each image $I_i \in \mathbb{R}^{3\times H\times W}$ is first patchified into $K$ tokens $t_I \in \mathbb{R}^{K\times C}$ using DINO feature extraction. These tokens are then processed through stacked transformer layers to yield frame-wise implicit geometric embeddings.

\noindent\textbf{RoomTour3D-IGR Architecture.}
As shown in Figure~\ref{fig:ext1}, RoomTour3D-IGR preserves the original components of RoomTour3D—namely, the scene encoder, large language model (LLM), and pose encoder for explicit geometry—while introducing the additional Spatial Encoder. The system dynamically selects between explicit and implicit geometry depending on data type. When processing simulator data, it behaves identically to the original RoomTour3D pipeline. When applied to web-based videos, the pose encoder is replaced by the Spatial Encoder, whose outputs are projected via a spatial projector to align with the LLM feature space. These implicit geometric embeddings are then fused with visual and textual representations to form a unified end-to-end navigation model capable of operating seamlessly across both simulator and real-world domains.

\section{Vision-and-Language Navigation Model}

In this section, we demonstrate how the proposed RoomTour3D data can be effectively utilized to train a generalist embodied agent. We first review the state-of-the-art large language model–based navigation framework, \textbf{NaviLLM}~\cite{navillm}. Then, we describe two RoomTour3D-specific training tasks: a \textbf{vision–instruction summarization task} for pretraining and an \textbf{action–instruction navigation task} for finetuning.

\subsection{Revisiting NaviLLM}

NaviLLM represents a state-of-the-art (SOTA) large language model (LLM) framework for embodied navigation, achieving superior results across major benchmarks such as CVDN and SOON. It processes panoramic inputs by encoding scene observations and integrating them with natural-language navigation instructions.  
Distinct tokens are used to differentiate input types: \textless hist\textgreater\ denotes historical observations, and \textless cand\textgreater\ represents candidate views at each decision step.

During training, NaviLLM takes as input a navigation instruction together with multiple candidate views. At each step, the model predicts the most appropriate view (action) from the candidates. The selected view is subsequently cached as a \textless hist\textgreater\ token, allowing the model to update its internal state for future reasoning.  
At the final step of the navigation sequence, all accumulated \textless hist\textgreater\ tokens are summarized as an auxiliary task to encourage comprehensive understanding of the agent’s traversed path.  
During inference, the model follows the same procedure: it sequentially accumulates historical observations (\textless hist\textgreater) and evaluates candidate views (\textless cand\textgreater) at each step to select the next optimal action. This iterative decision-making process ensures that the agent’s behavior remains consistent with both the observed environment and the given instruction.

\subsection{Summarization Task for Pretraining}

To exploit the rich sequential structure of RoomTour3D videos and enhance long-horizon reasoning, we adapt NaviLLM for a \textbf{summarization pretraining task}. Here, each frame within a trajectory is treated as a candidate view, wrapped by \textless cand\textgreater\ tokens. Similar to the panoramic nodes used in conventional VLN settings, these frames represent consecutive agent observations along a path.

As illustrated in Figure~\ref{fig:method}(a), we provide the LLM with both the sampled frame tokens and the corresponding task instruction in a unified prompt. The objective is to generate a trajectory summary describing object progression, spatial transitions, and room locations. Training follows a next-token prediction paradigm, consistent with standard autoregressive language model optimization.  
This pretraining stage allows the model to internalize spatial–temporal patterns and learn to reason about object relationships and scene continuity over extended sequences.

\subsection{Navigation Task for Finetuning}

To enable effective decision-making in realistic and large-scale environments, we further adapt NaviLLM for an \textbf{action–instruction finetuning task} based on the action-enriched trajectories from RoomTour3D. Unlike panoramic observations that represent a static viewpoint, our dataset provides consecutive frames from distinct locations and orientations, with only one frame leading toward the intended destination.

Each frame is again represented as a candidate action wrapped with \textless cand\textgreater\ tokens and is processed in the same format as panoramic views. As depicted in Figure~\ref{fig:method}(b), the model receives both the navigation instruction and historical frames, then selects the next optimal action from the available candidates. The chosen frame is cached as a \textless hist\textgreater\ token, which updates the model’s navigation history for subsequent decisions.

During finetuning, every frame corresponds to a potential navigable step. The next frame in the trajectory serves as the ground-truth target action, with \textless STOP\textgreater\ used as an alternative termination token. The model learns to iteratively predict the next step based on accumulated observations and the instruction context, effectively building an internal representation of the trajectory.  
At the final step, the model performs a summarization subtask that reinforces its ability to recall object transitions and room layouts, improving consistency between visual history and linguistic understanding. This hierarchical training approach strengthens the model’s capacity for long-term reasoning and precise action selection.

\vspace{-.8 em}
\subsection{Training with Implicit Geometry}

To integrate the proposed implicit geometry representations into the RoomTour3D training framework, we extend the model to the \textbf{RoomTour3D-IGR} configuration. We initialize this version with the pretrained weights from RoomTour3D and randomly initialize the projector module that bridges the Spatial Encoder and the LLM.  
The training corpus comprises both simulator-based data (e.g., Matterport3D) containing explicit geometric information and web-based videos containing only implicit geometry.

When training on simulator data, RoomTour3D-IGR behaves identically to the original RoomTour3D pipeline, using explicit geometric features such as distances and heading offsets. Conversely, when processing web-based video data, explicit geometry is replaced with implicit spatial embeddings extracted by the Spatial Encoder.  
During this phase, the Spatial Encoder remains frozen while the projector and LLM parameters are trainable. This ensures that the learned projector aligns the implicit spatial features with the LLM’s latent representation, enabling the joint system to generate meaningful geometric encodings.  
As a result, RoomTour3D-IGR effectively compensates for the absence of explicit 3D information in web videos and leverages a much larger portion of available data that would otherwise be discarded. This integration allows the agent to benefit from both high-quality simulator annotations and large-scale real-world video diversity, yielding a more robust and generalizable navigation policy.

\section{Tasks and Experiments}
\subsection{Benchmarks and Metrics}

\noindent\textbf{Benchmarks.}
During pretraining, we follow practice from NaviLLM~\cite{navillm} and perform teacher-forcing training on the combined dataset from our video-instruction data from RoomTour3D, together with CVDN~\cite{cvdn}, SOON~\cite{soon}, R2R~\cite{r2r}, REVERIE~\cite{reverie} and ScanQA~\cite{scanqa}, and augmented data from R2R and REVERIE. 
In the multi-task fine-tuning stage, we alternate between teacher forcing and student forcing on the combined data from our action-instruction data from RoomTour3D, together with CVDN, SOON, R2R, REVERIE, ScanQA and LLaVA-23k~\cite{liu2023llava}.

To evaluate the impact of our data on navigation agent training, we test on CVDN, SOON, R2R, and REVERIE. CVDN requires navigating towards a target by understanding dialog history, linking dialogue comprehension to actions. 
SOON tasks the agent with locating objects without bounding boxes, emphasizing semantic-visual alignment. 
R2R involves following step-by-step instructions, requiring dynamic progress tracking and precise alignment with navigational history. 
REVERIE focuses on localizing distant objects based on concise instructions, aided by ground truth bounding boxes at waypoints.

\begin{table*}[htp!]
  \caption{Overall comparison with the baseline methods. Our RoomTour3D data can boost NaviLLM by a margin on SOON, R2R and REVERIE on SPL metric and on CVDN GP metric. $^\star$ denotes reproduced results. $^{\dagger}$ denotes involving all RoomTour3D data into geometry-aware training, refering to Table~\ref{table:igr_single} for detailed comparison.}
  \label{tab:supervised_table}
  \centering
  \footnotesize
  \setlength{\tabcolsep}{10pt}
  \resizebox{0.95\linewidth}{!}{%
  \begin{tabular}{lcccccccc}
    \toprule
    \multirow{2}{*}{\textbf{Methods}} &
    \multicolumn{2}{c}{\textbf{CVDN}} &
    \multicolumn{2}{c}{\textbf{SOON}} &
    \multicolumn{2}{c}{\textbf{R2R}} &
    \multicolumn{2}{c}{\textbf{REVERIE}} \\
    \cmidrule(lr){2-3}\cmidrule(lr){4-5}\cmidrule(lr){6-7}\cmidrule(lr){8-9}
     & Val-U & Test & Val-U & Test & Val-U & Test & Val-U & Test \\
    \midrule
    \multicolumn{9}{l}{\textit{Models Focusing on Single Task}} \\
        \midrule
    \addlinespace[2pt]
    PREVALENT~\cite{hao2020towards}  & 3.15 & 2.44 & -    & -    & 53 & 51 & -    & - \\
    HOP~\cite{qiao2022hop}           & 4.41 & 3.24 & -    & -    & 57 & 59 & 26.1 & 24.3 \\
    HAMT~\cite{hamt}                 & 5.13 & 5.58 & -    & -    & 61 & 60 & 30.2 & 26.7 \\
    DUET~\cite{duet}                 &  -   &  -   & 22.6 & 21.4 & 60 & 58 & 33.7 & 36.0 \\
    VLN-SIG~\cite{li2023improving}   & 5.52 & 5.83 & -    & -    & 62 & 60 & -    & - \\
    VLN-PETL~\cite{qiao2023vln}      & 5.69 & 6.13 & -    & -    & 60 & 58 & 27.7 & 26.7 \\
    DiscussNav~\cite{long2024discuss} &-&-&-&-&40&-&-&-\\
    NavGPT2~\cite{zhou2025navgpt}    &  -   &  -   & -    & -    & 61 & 60 & -    & - \\
    AZHP ~\cite{gao2023adaptive}&-&-&26.6&-&61&60&36.6&35.8\\
    BEV-BERT~\cite{an2023bevbert}    &  -   &  -   & -    & -    & 64 & 60 & 36.4 & 36.4 \\
    \midrule
    \multicolumn{9}{l}{\textit{Unified Model For All Tasks}} \\
        \midrule
    \addlinespace[2pt]
    NaviLLM (w. Pretrain)~\cite{navillm}       & 6.16 & 7.90 & 29.2 & 26.3 & 59 & 60 & 35.7 & 32.3 \\
    NaviLLM (w. Pretrain)$^\star$              & 6.09 &  -   & 28.0 &  -   & 57 &  -  & 31.4 &  -   \\
    $\textbf{RoomTour3D--IGR (Ours)}^{\dagger}$&  7.38   &  8.06   &  32.7   &  29.4   &  66 &  64  &   39.2  &  37.0   \\
    \bottomrule
  \end{tabular}
  }
  \vskip -0.05in
\end{table*}

\noindent\textbf{Evaluation Metrics.}
For the navigation tasks, we follow the evaluation methodology from \cite{r2r} using the following navigation metrics: \textbf{Success Rate (SR)}, which measures whether the agent reaches the target location within a set distance threshold; \textbf{Success Rate Weighted by Path Length (SPL)}, which is the SR adjusted by the ratio of the ground truth path length to the actual path traveled; \textbf{Goal Progress (GP)}, the advancement in meters towards the goal. GP is utilized for the CVDN dataset, whereas SR and SPL are the metrics for other datasets.

\subsection{Implementation details}
\label{sup:impl_det}
Following the practice from NaviLLM~\cite{navillm}, we fine-tune the multi-view fusion module and the LLM. The multi-view fusion module consists of a 2-layer transformer encoder with a hidden size of 1024, and the LLM is built upon Vicuna-7B-v1.1~\cite{vicuna2023}. The ViT in the scene encoder is EVA-CLIP-02-Large, which remains frozen during training. Our Spatial Encoder is built upon the pre-trained VGGT~\cite{wang2025vggt} without the prediction heads. The Spatial Encoder also remains frozen during training. 
Our RoomTour3D training follows a two-stage strategy using the Adam optimizer with a learning rate of 3e-5. The model is trained for 2500 steps in the pre-training stage and 1250 steps in the multi-task fine-tuning stage, with a batch size of 256. Our RoomTour3D--IGR training is built upon RoomTour3D with a learning rate of 5e-6 and 5000 steps using multi-task learning. The training process utilizes 4$\times$8 Nvidia A100 GPUs. 
During testing, we employ a sampling strategy with a temperature of 0.01 for the SOON and REVERIE tasks to encourage exploration, while a greedy strategy is used for other tasks. This approach ensures robust performance across various evaluation scenarios.

\subsection{Comparison on Supervised Tasks}

As shown in Table~\ref{tab:supervised_table}, we performed a one-time fine-tuning on the four tasks in a fully supervised manner. To begin, our experiments reiterate the superiority of multitask training over single-task training. Also, incorporating our RoomTour3D data into the pre-training process led to consistent improvements across all metrics on Val-U, achieving state-of-the-art results in the GP metric in the CVDN dataset. While the improvement on the CVDN datasets is modest, the most significant boost compared to the reproduced baseline is observed in R2R Val-U, REVERIE Val-U and SOON Val-U, with gains of approximately 7\%, 3.5\% and 3.5\%, respectively. It is also surprising to witness that our RoomTour3D-IGR also achieves steady improvement on the held-out testing set. 
The improvement in R2R is largely driven by enhanced spatial awareness, stemming from the inclusion of proximity data, which helps the model better understand object distance and position. Similarly, gains in REVERIE are attributed to a combination of open-vocabulary tags, spatial awareness, and the addition of room type data, which encourages the model to infer the layout of environments, thereby boosting its spatial reasoning capabilities. Moreover, our use of open-ended instructions allows the model to adapt flexibly to diverse scenarios, fostering more robust and generalizable performance and better contextual understanding.

\begin{table*}[t]
  \centering
\caption{RoomTour3D$_{\text{Desc}}$ and RoomTour3D$_{\text{Action}}$ stand for description-enriched trajectories only and action-enriched trajectories. Involving description-enriched trajectories and action-enriched trajectories achieves improvement over multiple benchmarks. The inclusion of implicit geometric representation achieves a further performance boost on both Val-U and Test sets. (E) denotes explicit geometric trajectories, and (I) denotes implicit geometry representation.}
\vspace{-0.2cm}
  \setlength{\tabcolsep}{10pt}
\resizebox{0.95\linewidth}{!}{%
\begin{tabular}{lccccccccccc}
    \toprule
    \multirow{2}{*}{\textbf{Methods}} &
    \multicolumn{2}{c}{\textbf{Data type}} &
    \multicolumn{2}{c}{\textbf{CVDN}} &
    \multicolumn{2}{c}{\textbf{SOON}} &
    \multicolumn{2}{c}{\textbf{R2R}} &
    \multicolumn{2}{c}{\textbf{REVERIE}} \\
    \cmidrule(lr){2-3}\cmidrule(lr){4-5}\cmidrule(lr){6-7}\cmidrule(lr){8-9}\cmidrule(lr){10-11}
     & Desc & Action & Val-U & Test & Val-U & Test & Val-U & Test & Val-U & Test \\
    \midrule
    NaviLLM (w. Pretrain)$^\star$ 
        & $\times$ & $\times$ 
        & 6.09 &  -   & 28.0 &  -   & 57 &  -  & 31.4 &  -   \\
    RoomTour3D$_{\text{Desc}}$   
        & $\checkmark$ & $\times$ 
        & 6.96 & 7.55 & 30.2 & 26.5 & 62 & 62 & 37.1 & 35.1 \\
    \hline
    RoomTour3D$_{\text{Action}}$   
        & $\checkmark$ & $\checkmark$ (E) 
        & 6.33 & 7.22 & 31.7 & 27.8 & 62 & 62 & 37.4 & 36.4 \\
    RoomTour3D--IGR 
        & $\checkmark$ & $\checkmark$ (E+I) 
        & 7.10 & 7.73 & 32.0 & 28.3 & 64 & 63 & 38.0 & 36.5 \\
    \bottomrule
\end{tabular}%
}
\label{tab:geo_data_training}

  \vskip -0.1in
\end{table*}

\begin{table*}[t!]
  \caption{Ablation study on the input modalities for the trajectory summarization task.}
  \vskip -0.1in
  \centering
  \footnotesize
    \setlength{\tabcolsep}{10pt}
    \resizebox{0.95\linewidth}{!}{%
  \begin{tabular}{ccc ccccccc}
    \toprule
    \multirow{2}{*}{\textbf{Object tags}} & 
    \multirow{2}{*}{\textbf{Depth / Box}} & 
    \multirow{2}{*}{\textbf{Room type}} & 
    \multicolumn{1}{c}{\textbf{CVDN}} &
    \multicolumn{2}{c}{\textbf{SOON}} &
    \multicolumn{2}{c}{\textbf{R2R}} &
    \multicolumn{2}{c}{\textbf{REVERIE}} \\
    \cmidrule(lr){4-4}\cmidrule(lr){5-6}\cmidrule(lr){7-8}\cmidrule(lr){9-10}
     & & & GP$\uparrow$ & SR$\uparrow$ & SPL$\uparrow$ & SR$\uparrow$ & SPL$\uparrow$ & SR$\uparrow$ & SPL$\uparrow$ \\
    \midrule
    $\times$ & $\times$ & $\times$ & 6.09 & 33.6 & 28.0 & 66 & 57 & 38.3 & 31.4 \\
    \checkmark & $\times$ & $\times$ & 5.41 & 32.5 & 26.5 & 64 & 56 & 42.5 & 34.4 \\
    \checkmark & \checkmark & $\times$ & 6.49 & 37.6 & \textbf{30.4} & 68 & 62 & 41.7 & 36.0 \\
    \checkmark & \checkmark & \checkmark & \textbf{6.96} & \textbf{38.8} & 30.2 & \textbf{69} & \textbf{62} & \textbf{43.3} & \textbf{37.1} \\
    \bottomrule
  \end{tabular}
  }
  \label{tab:ablation}
  \vskip -0.1in
\end{table*}

\begin{table}[t!]
  \caption{Overall comparison with SOTA zero-shot methods on R2R. $^{\dagger}$ denotes training exclusive navigable actions. $^{\star}$ denotes using 36 views setting.}
  \centering
  \footnotesize
  \setlength{\tabcolsep}{12pt}
  \resizebox{0.9\linewidth}{!}{%
  \begin{tabular}{lcc}
    \toprule
    \textbf{Methods} & SR$\uparrow$ & SPL$\uparrow$ \\
    \midrule
    Random Walk~\cite{pan2023langnav} & 3 & 2 \\
    \midrule
    \multicolumn{3}{l}{\textit{Commercial Models}} \\
    \addlinespace[2pt]
        \midrule
    NavGPT (GPT-3.5)$^{\star}$~\cite{zhou2024navgpt} & 13.89 & 9.12 \\
    NavGPT (GPT-4)~\cite{zhou2024navgpt} & 34 & 29 \\
    MapGPT (GPT-4)~\cite{chen2024mapgpt} & 38.8 & 25.8 \\
    MapGPT (GPT-4V)~\cite{chen2024mapgpt} & \textbf{43.7} & 34.8 \\
    DiscussNav (GPT-4)~\cite{long2023discuss} & 43 & \textbf{40} \\
    \midrule
    \multicolumn{3}{l}{\textit{Open-source Models}} \\
    \addlinespace[2pt]
        \midrule
    LangNav (LLaMA2-7B)~\cite{pan2023langnav} & 0 & 0 \\
    NavCoT (LLaMA2-7B)~\cite{lin2024navcot} & 7.78 & 6.50 \\
    DuET (Init. LXMERT~\cite{tan2019lxmert}) & 1 & 0 \\
    NaviLLM$^{\dagger}$~\cite{navillm} & 0 & 0 \\
    RoomTour3D (Ours) & 14.33 & 10.86 \\
    RoomTour3D–IGR (Ours) &19.21&14.60\\
    \bottomrule
  \end{tabular}
  }
  \label{tab:zero_shot}
\end{table}

\subsection{Comparison on Zero-shot Task}

To further demonstrate the substantial indoor knowledge contained in our data and its effectiveness for embodied action and language instructions, we conduct zero-shot experiments on embodied action prediction, as shown in Table~\ref{tab:zero_shot}. 

We removed all action and geometric data from the training datasets and retrained NaviLLM with and without our RoomTour3D dataset. Without action prediction data, NaviLLM lacked the ability to learn effective navigable action selection. However, with the inclusion of our action-enriched trajectories, NaviLLM achieved a 14.33\% SR and a 10.86\% SPL, outperforming open-source models built on LLaMA-7B and reaching results comparable to NavGPT~\cite{zhou2024navgpt}, which leverages GPT-3.5. These improvements validate the effectiveness of our 3D trajectories mined from room tour video reconstructions and emphasize the value of our action-enriched trajectories. 
Compared to using only the explicit geometry encoding from RoomTour3D, the additional use of implicit geometry encoding in RoomTour3D–IGR achieved even higher performance, demonstrating a clear improvement. This gain is likely due to the implicit geometry representation’s ability to expand the effective training dataset and provide the navigation agent with richer prior knowledge. Together, they highlight the substantial contribution of our dataset in advancing open-world navigation research.

\subsection{Ablation study}

\noindent\textbf{Effect of different RoomTour3D trajectories.}
Table~\ref{tab:geo_data_training} compares the NaviLLM baseline with models trained on description-enriched trajectories and action-enriched trajectories. Using only description-enriched trajectories (RoomTour3D\textsubscript{Desc}) already improves the baseline across all benchmarks—for example, SOON Val-U increases from 28.0 to 30.2, R2R Val-U improves from 57 to 62 (with 62 on Test), and REVERIE Val-U rises from 31.4 to 37.1 (35.1 on Test). Incorporating action-enriched trajectories with explicit geometry (RoomTour3D\textsubscript{Action}) yields further gains, achieving 31.7 on SOON and 37.4 on REVERIE, while maintaining performance on R2R and CVDN. Finally, adding implicit geometry representation (\textbf{RoomTour3D--IGR}) delivers the best overall results, 7.10 on CVDN, 32.0 on SOON, 64 on R2R, and 38.0 on REVERIE, demonstrating that both description- and action-enriched trajectories are beneficial and that implicit geometry further enhances generalization. Note that within this range of experiments, only action-enriched trajectories are used for geometry training.

\noindent\textbf{Effect of open-world semantics and spatial awareness.}
As shown in Table~\ref{tab:ablation}, we analyzed the impact of various information types on instruction generation. Adding object variety significantly improved performance on REVERIE with increase SPL from 31.35\% to 34.37\%, as this dataset relies on object grounding. However, it had no direct impact on SOON, possibly because SOON relies solely on detailed textual descriptions without explicit bounding box annotations. After introducing depth estimation, which helps determine the relative distances of objects, the performance on SOON, R2R and REVERIE achieve marginal boosts. This demonstrates that enhancing spatial awareness significantly contributes to indoor navigation tasks. Furthermore, incorporating room locations, which capture the scene semantics along the trajectory, provided a moderate boost across all four VLN tasks. This further highlights the critical role of object variety and spatial awareness in improving navigation performance.

\begin{table}[t]
  \caption{Ablation study on geometric encoding. Compared with using only explicit or only implicit encoding, our RoomTour3D–IGR combines both and serves as a better choice. We report SPL for R2R, REVERIE and SOON, GP for CVDN on Val-U.}
  \centering
  \small
\resizebox{0.95\linewidth}{!}{%
  \begin{tabular}{lcccc}
    \toprule
    \textbf{Geometry encoding} & \textbf{CVDN} & \textbf{SOON} & \textbf{R2R} & \textbf{REVERIE} \\
    \midrule
    Explicit            & 6.33 & 31.7 & 62 & 37.4 \\
    Implicit           & 7.07 & 30.9 & 64 & 36.8 \\
    Explicit + Implicit      & 7.38 & 32.7 & 66 & 39.2 \\
    \bottomrule
  \end{tabular}
  }
  \label{tab:ab_igr}
  \vskip -0.1in
\end{table}

\begin{table}[tbp!]
\centering
\caption{Ablation on training data usage. As implicit geometry relaxes reconstruction constraints, more trajectories can be included for geometric training. Consistent performance improvement across CVDN, SOON, R2R, and REVERIE (GP reported for CVDN, SPL for others) can be observed.$\dagger$ means involvement of all RoomTour3D data.}
\vspace{-0.2cm}
\resizebox{0.95\linewidth}{!}{%
\begin{tabular}{lcccc}
    \toprule
    \textbf{Methods} & \textbf{CVDN} & \textbf{SOON} & \textbf{R2R} & \textbf{REVERIE} \\
    \midrule
     RoomTour3D$_{\text{Desc}}$   
        &  6.96 & 30.2& 62  & 37.1  \\
    \textbf{RoomTour3D--IGR  }      & 7.10 & 32.0 & 64 & 38.0 \\
    $\textbf{RoomTour3D--IGR  }^{\dagger}$ & 7.38 & 32.7 & 66 & 39.2 \\
    \bottomrule
\end{tabular}%
}
\label{table:igr_single}
\end{table}

\noindent\textbf{Impact of geometry training scale.} 
Table~\ref{table:igr_single} highlights how implicit geometry allows more data to be exploited for geometric training. Starting from description-enriched trajectories, adding implicit geometry already improves performance (e.g., +0.14 on CVDN and +0.9 on REVERIE), confirming its effectiveness even without enlarging the geometry training set (\ie, the action-enriched trajectories). When \emph{all} trajectories are included with implicit geometric encoding, the performance further improves from 64 to 66 on R2R, and from 38.0 to 39.2 on REVERIE, from 7.10 to 7.38 on CVDN, from 32.0 to 32.7 on SOON. This demonstrates that implicit representations not only bypass the strict filtering of explicit reconstruction but also unlock a much larger portion of training data, which translates into consistent and measurable improvements across diverse navigation tasks.

\noindent\textbf{Source of geometric information.} Table~\ref{tab:ab_igr} presents the ablation study on different sources of geometric encoding. Using only explicit geometry achieves a baseline of 6.33 on CVDN, 31.7 on SOON, 62 on R2R, and 37.4 on REVERIE. In contrast, implicit geometry encoding provides stronger generalization in most benchmarks, particularly on CVDN (7.07 vs. 6.33) and R2R (64 vs. 62), although it slightly underperforms on REVERIE. When combining both explicit and implicit geometry (RoomTour3D--IGR), the model achieves the best overall performance across all datasets—7.38 on CVDN, 32.7 on SOON, 66 on R2R, and 39.2 on REVERIE. These results confirm that implicit encoding is more robust than explicit alone, while the integration of both sources offers complementary benefits and leads to the strongest performance.

\subsection{Robustness to Visual Degradations}
\begin{figure*}[t]
    \centering
    \includegraphics[width=0.98\linewidth]{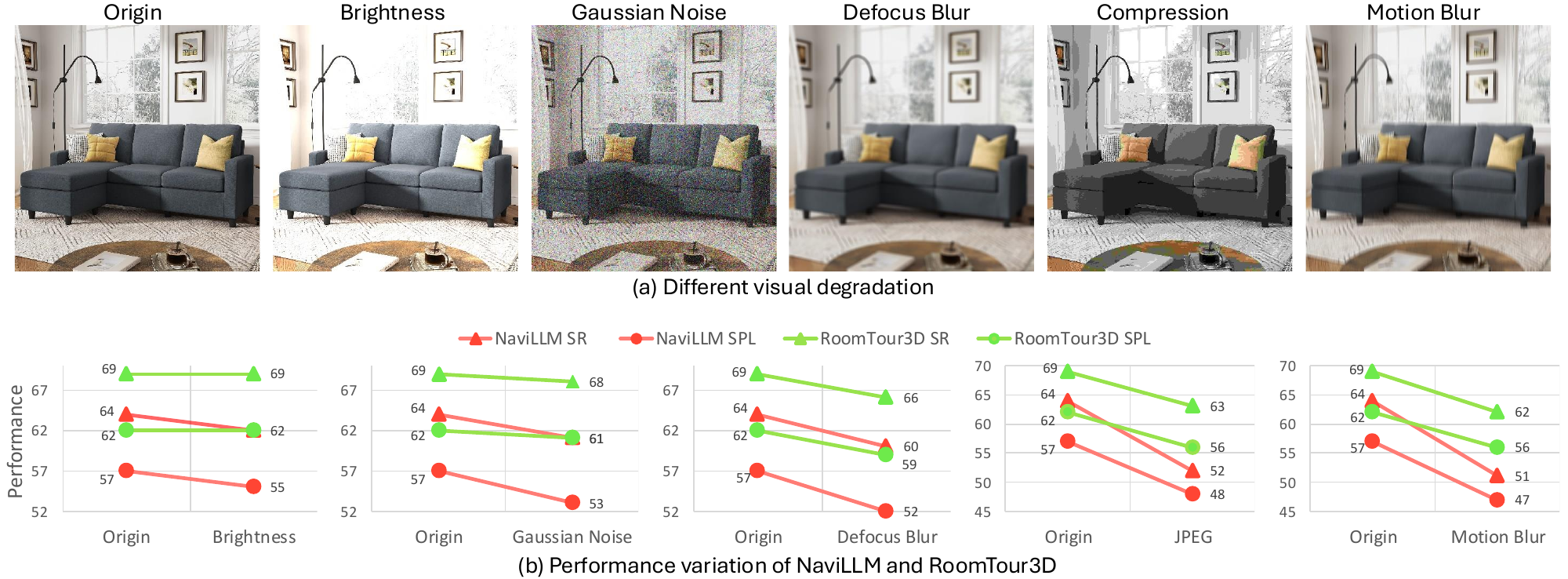}
    \vskip -0.1in
    \caption{Visual robustness evaluation under common degradations on R2R Val Unseen. 
The figure illustrates four types of perturbations (Gaussian noise, motion blur, JPEG compression, defocus blur, and brightness) with image examples, together with SPL and SR results. 
Compared to NaviLLM, our RoomTour3D-trained agent suffers smaller performance drops under all degradations, highlighting improved tolerance in real-world. }
    \label{fig:exp_ext}
        \vskip -0.15in
\end{figure*}

\begin{figure*}[!ht]
    \centering
    \includegraphics[width=0.92\linewidth]{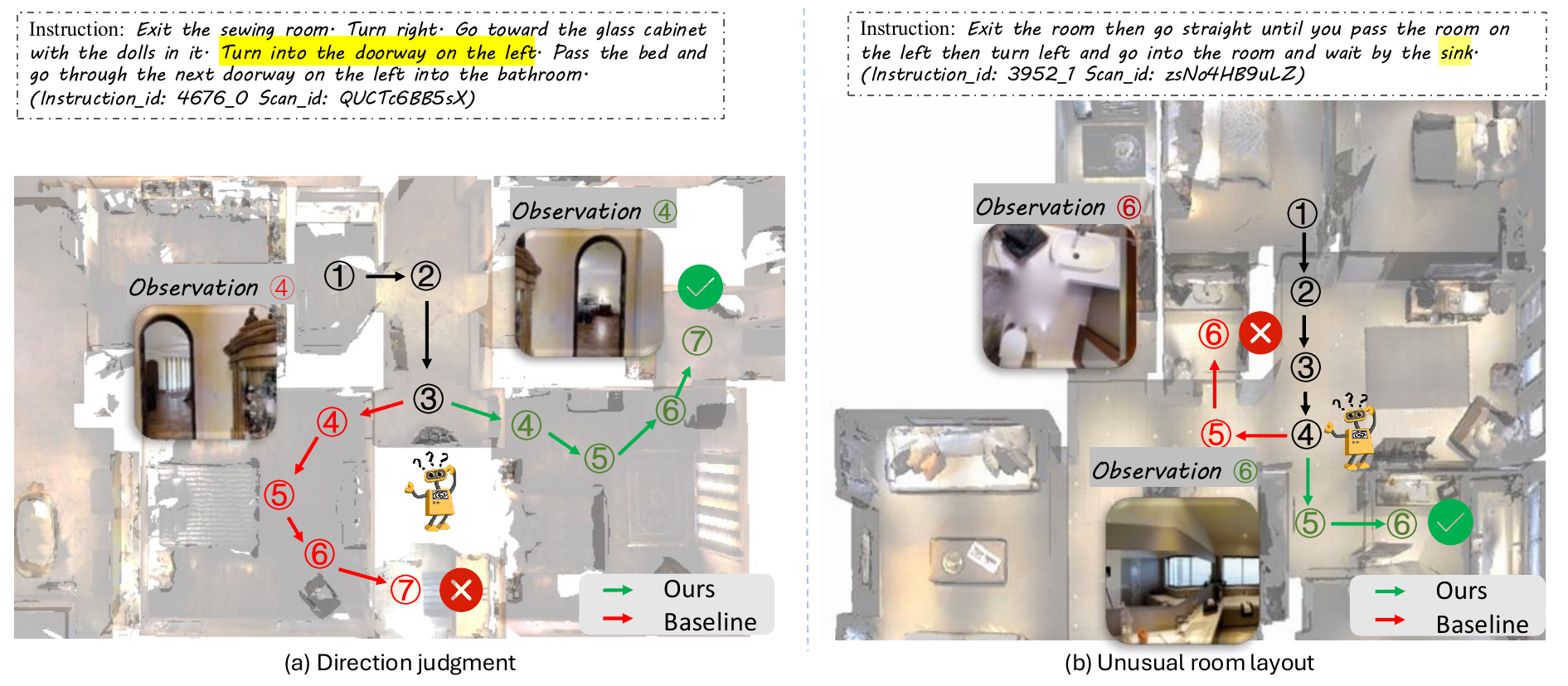}
        \vskip -0.1in
    \caption{Visualization of navigation paths of NaviLLM~\cite{navillm} and our RoomTour3D model on R2R-unseen. \textcircled{1} denotes the starting location. The red circle marks an incorrect endpoint, while the green circle indicates the correct one. (a) In a scenario requiring directional judgment, our method makes the correct decision, whereas the baseline is misled by a similar-looking entrance at the waypoint and mistakenly turns right. (b) In an unusual room layout, where two different sinks appear in close proximity, the baseline is distracted by the first incorrect sink it encounters. In contrast, our method follows the instruction “go straight until you pass” more reliably and successfully reaches the correct sink.}
    \label{fig:visualization}
\end{figure*}

While the original NaviLLM model was trained on simulator-rendered data, such data are inherently clean and free of noise, which diverges from the visual inputs a real navigation robot would encounter. In practice, robots may experience camera shake during movement or out-of-focus frames, leading to degraded observations that significantly affect performance. Our RoomTour3D dataset, being web-based, is closer to real-world conditions: human operators who captured these videos naturally introduced motion jitter, defocus, and other imperfections. We therefore hypothesize that training on RoomTour3D enhances the model’s robustness and tolerance to real-world disturbances. Following prior work \cite{hendrycks2019benchmarking}, we systematically introduce five common degradation types: Gaussian Noise (simulating sensor noise), Motion Blur (simulating motion-induced shake), JPEG (corrupted by JPEG compression), Defocus Blur (simulating camera defocus), and Brightness change (simulating camera over-exposure) to evaluate the visual robustness of our approach.

Compared with NaviLLM, our RoomTour3D model demonstrates stronger robustness against visual degradations. As shown in Figure~\ref{fig:exp_ext}, both models experience performance drops under degraded inputs, but the impact on RoomTour3D is consistently smaller. In mild cases such as Gaussian Noise and Brightness changes, RoomTour3D suffers almost no degradation, whereas NaviLLM shows around a 3\% drop in SPL and SR. In more severe scenarios such as Motion Blur, NaviLLM encounters a substantial 11\% decline, while RoomTour3D shows only about a 7\% decrease. This advantage arises because our RoomTour3D dataset, being web-based, inherently captures real-world imperfections, such as motion jitter, defocus, and compression artifacts, introduced during human video recording. Training on RoomTour3D therefore equips the model with greater tolerance to real-world disturbances and enhances its overall robustness.

\subsection{Navigation Case Visualization}
As shown in Figure~\ref{fig:visualization}, we highlight two representative cases where our RoomTour3D model improves upon NaviLLM~\cite{navillm}. In case (a), at \textcircled{3}, the agent is required to make a left turn following the instruction \textit{“Turn into the doorway on the left.”} However, the baseline incorrectly turns right at the designated decision point, causing it to deviate from the intended path. Once this mistake occurs, even with scene-graph history, the model struggles to realign with the correct trajectory. In contrast, training with RoomTour3D provides richer spatial and semantic supervision, enabling the agent to make more accurate decisions in such directional judgment scenarios.

In case (b), at \textcircled{4}, the agent must follow the instruction \textit{“Go straight until you pass the room”} to reach the correct sink among two sinks located in close proximity. The baseline, however, is distracted by the first sink it encounters and deviates onto an incorrect path. This unusual room layout, where multiple similar objects appear nearby, causes the baseline to overlook the critical instruction phrase “go straight until.” By leveraging RoomTour3D, which offers richer priors about room layouts and spatial semantics, our method better adheres to the instruction and successfully navigates to the correct sink.

\section{Conclusion}

In this paper, we presented RoomTour3D, a novel dataset automatically constructed from large collections of room tour videos for vision-and-language navigation (VLN). By leveraging the sequential continuity of video streams and integrating both object diversity and spatial awareness, our framework generates over 200K navigation instructions and 17K action-enriched trajectories across 1,847 indoor scenes. From these videos, we further extract navigable trajectories based on frame sequences and reconstructed 3D environments, leading to significant performance improvements and new state-of-the-art results on the SOON and REVERIE benchmarks. Moreover, this scalable pipeline enables the training of a zero-shot navigation agent, underscoring the effectiveness and generalizability of RoomTour3D for advancing embodied VLN research.

Beyond explicit geometry, we extend the framework with implicit geometry representations (RoomTour3D-IGR), which infer spatial cues directly from RGB imagery without depending on fragile 3D reconstruction. This enhancement mitigates the high failure rate of reconstruction in unconstrained web videos and greatly increases the usable data volume. Experimental results confirm that implicit geometry not only substitutes explicit geometry effectively but also produces consistent gains across multiple VLN benchmarks, improving both robustness and scalability. Collectively, these contributions establish RoomTour3D and its implicit variant as powerful resources for developing next-generation embodied navigation systems capable of learning from large-scale, real-world visual data.

    \bibliographystyle{IEEEtran}
    \bibliography{main}

@String(CVPR= {IEEE Conf. Comput. Vis. Pattern Recog.})

@String(ICCV= {Int. Conf. Comput. Vis.})

@String(ECCV= {Eur. Conf. Comput. Vis.})

@String(TOG= {ACM Trans. Graph.})

@String(AAAI = {AAAI})

@String(CVPR  = {CVPR})

@String(ICCV  = {ICCV})

@String(ECCV  = {ECCV})

@String(TOG   = {ACM TOG})

@misc{navillm,
      title={Towards Learning a Generalist Model for Embodied Navigation}, 
      author={Duo Zheng and Shijia Huang and Lin Zhao and Yiwu Zhong and Liwei Wang},
      year={2023},
      eprint={2312.02010},
      archivePrefix={arXiv},
      primaryClass={cs.CV}
}

@misc{youtube_vln,
      title={Learning Vision-and-Language Navigation from YouTube Videos}, 
      author={Kunyang Lin and Peihao Chen and Diwei Huang and Thomas H. Li and Mingkui Tan and Chuang Gan},
      year={2023},
      eprint={2307.11984},
      archivePrefix={arXiv},
      primaryClass={cs.CV}
}

@misc{airbert,
      title={Airbert: In-domain Pretraining for Vision-and-Language Navigation}, 
      author={Pierre-Louis Guhur and Makarand Tapaswi and Shizhe Chen and Ivan Laptev and Cordelia Schmid},
      year={2021},
      eprint={2108.09105},
      archivePrefix={arXiv},
      primaryClass={cs.CV}
}

@InProceedings{vln_bert,
    author    = {Hong, Yicong and Wu, Qi and Qi, Yuankai and Rodriguez-Opazo, Cristian and Gould, Stephen},
    title     = {A Recurrent Vision-and-Language {BERT} for Navigation},
    booktitle = {Proceedings of the IEEE/CVF Conference on Computer Vision and Pattern Recognition (CVPR)},
    month     = {June},
    year      = {2021},
    pages     = {1643-1653}
}

@misc{r2r,
      title={Vision-and-Language Navigation: Interpreting visually-grounded navigation instructions in real environments}, 
      author={Peter Anderson and Qi Wu and Damien Teney and Jake Bruce and Mark Johnson and Niko Sünderhauf and Ian Reid and Stephen Gould and Anton van den Hengel},
      year={2018},
      eprint={1711.07280},
      archivePrefix={arXiv},
      primaryClass={cs.CV}
}

@misc{cvdn,
      title={Vision-and-Dialog Navigation}, 
      author={Jesse Thomason and Michael Murray and Maya Cakmak and Luke Zettlemoyer},
      year={2019},
      eprint={1907.04957},
      archivePrefix={arXiv},
      primaryClass={cs.CL}
}

@misc{soon,
      title={{SOON}: Scenario Oriented Object Navigation with Graph-based Exploration}, 
      author={Fengda Zhu and Xiwen Liang and Yi Zhu and Xiaojun Chang and Xiaodan Liang},
      year={2021},
      eprint={2103.17138},
      archivePrefix={arXiv},
      primaryClass={cs.CV}
}

@misc{reverie,
      title={{REVERIE}: Remote Embodied Visual Referring Expression in Real Indoor Environments}, 
      author={Yuankai Qi and Qi Wu and Peter Anderson and Xin Wang and William Yang Wang and Chunhua Shen and Anton van den Hengel},
      year={2020},
      eprint={1904.10151},
      archivePrefix={arXiv},
      primaryClass={cs.CV}
}

@misc{eqa_1,
      title={Embodied Question Answering in Photorealistic Environments with Point Cloud Perception}, 
      author={Erik Wijmans and Samyak Datta and Oleksandr Maksymets and Abhishek Das and Georgia Gkioxari and Stefan Lee and Irfan Essa and Devi Parikh and Dhruv Batra},
      year={2019},
      eprint={1904.03461},
      archivePrefix={arXiv},
      primaryClass={cs.CV}
}

@misc{eqa_2,
      title={Embodied Question Answering}, 
      author={Abhishek Das and Samyak Datta and Georgia Gkioxari and Stefan Lee and Devi Parikh and Dhruv Batra},
      year={2017},
      eprint={1711.11543},
      archivePrefix={arXiv},
      primaryClass={cs.CV}
}

@misc{new_path,
      title={A New Path: Scaling Vision-and-Language Navigation with Synthetic Instructions and Imitation Learning}, 
      author={Aishwarya Kamath and Peter Anderson and Su Wang and Jing Yu Koh and Alexander Ku and Austin Waters and Yinfei Yang and Jason Baldridge and Zarana Parekh},
      year={2023},
      eprint={2210.03112},
      archivePrefix={arXiv},
      primaryClass={cs.LG}
}

@misc{hm3d_auto,
      title={Learning from Unlabeled 3D Environments for Vision-and-Language Navigation}, 
      author={Shizhe Chen and Pierre-Louis Guhur and Makarand Tapaswi and Cordelia Schmid and Ivan Laptev},
      year={2022},
      eprint={2208.11781},
      archivePrefix={arXiv},
      primaryClass={cs.CV}
}

@InProceedings{scalevln,
      author    = {Zun Wang and Jialu Li and Yicong Hong and Yi Wang and Qi Wu and Mohit Bansal and Stephen Gould and Hao Tan and Yu Qiao},
      title     = {Scaling Data Generation in Vision-and-Language Navigation},
      booktitle = {ICCV 2023},
      year      = {2023}
  }

@misc{self_explore1,
      title={Less is More: Generating Grounded Navigation Instructions from Landmarks}, 
      author={Su Wang and Ceslee Montgomery and Jordi Orbay and Vighnesh Birodkar and Aleksandra Faust and Izzeddin Gur and Natasha Jaques and Austin Waters and Jason Baldridge and Peter Anderson},
      year={2022},
      eprint={2111.12872},
      archivePrefix={arXiv},
      primaryClass={cs.CV}
}

@misc{self_explore2,
      title={Visual-Language Navigation Pretraining via Prompt-based Environmental Self-exploration}, 
      author={Xiwen Liang and Fengda Zhu and Lingling Li and Hang Xu and Xiaodan Liang},
      year={2022},
      eprint={2203.04006},
      archivePrefix={arXiv},
      primaryClass={cs.CV}
}

@misc{aug1,
      title={Learning to Navigate Unseen Environments: Back Translation with Environmental Dropout}, 
      author={Hao Tan and Licheng Yu and Mohit Bansal},
      year={2019},
      eprint={1904.04195},
      archivePrefix={arXiv},
      primaryClass={cs.CL}
}

@misc{aug2,
      title={Speaker-Follower Models for Vision-and-Language Navigation}, 
      author={Daniel Fried and Ronghang Hu and Volkan Cirik and Anna Rohrbach and Jacob Andreas and Louis-Philippe Morency and Taylor Berg-Kirkpatrick and Kate Saenko and Dan Klein and Trevor Darrell},
      year={2018},
      eprint={1806.02724},
      archivePrefix={arXiv},
      primaryClass={cs.CV}
}

@misc{aug3,
      title={{EnvEdit}: Environment Editing for Vision-and-Language Navigation}, 
      author={Jialu Li and Hao Tan and Mohit Bansal},
      year={2022},
      eprint={2203.15685},
      archivePrefix={arXiv},
      primaryClass={cs.CV}
}

@misc{aug5,
      title={Counterfactual Vision-and-Language Navigation via Adversarial Path Sampling}, 
      author={Tsu-Jui Fu and Xin Eric Wang and Matthew Peterson and Scott Grafton and Miguel Eckstein and William Yang Wang},
      year={2020},
      eprint={1911.07308},
      archivePrefix={arXiv},
      primaryClass={cs.CV}
}

@misc{aug4,
      title={Pathdreamer: A World Model for Indoor Navigation}, 
      author={Jing Yu Koh and Honglak Lee and Yinfei Yang and Jason Baldridge and Peter Anderson},
      year={2021},
      eprint={2105.08756},
      archivePrefix={arXiv},
      primaryClass={cs.CV}
}

@misc{aug6,
      title={Improving Cross-Modal Alignment in Vision Language Navigation via Syntactic Information}, 
      author={Jialu Li and Hao Tan and Mohit Bansal},
      year={2021},
      eprint={2104.09580},
      archivePrefix={arXiv},
      primaryClass={cs.CL}
}

@misc{aug7,
      title={Vision-Language Navigation with Random Environmental Mixup}, 
      author={Chong Liu and Fengda Zhu and Xiaojun Chang and Xiaodan Liang and Zongyuan Ge and Yi-Dong Shen},
      year={2021},
      eprint={2106.07876},
      archivePrefix={arXiv},
      primaryClass={cs.CV}
}

@InProceedings{hamt,
author       = {Chen, Shizhe and Guhur, Pierre-Louis and Schmid, Cordelia and Laptev, Ivan},
title        = {History Aware multimodal Transformer for Vision-and-Language Navigation},
booktitle    = {NeurIPS},
year         = {2021},
}

@misc{duet,
      title={Think Global, Act Local: Dual-scale Graph Transformer for Vision-and-Language Navigation}, 
      author={Shizhe Chen and Pierre-Louis Guhur and Makarand Tapaswi and Cordelia Schmid and Ivan Laptev},
      year={2022},
      eprint={2202.11742},
      archivePrefix={arXiv},
      primaryClass={cs.CV}
}

@misc{metaexplore,
      title={Meta-Explore: Exploratory Hierarchical Vision-and-Language Navigation Using Scene Object Spectrum Grounding}, 
      author={Minyoung Hwang and Jaeyeon Jeong and Minsoo Kim and Yoonseon Oh and Songhwai Oh},
      year={2023},
      eprint={2303.04077},
      archivePrefix={arXiv},
      primaryClass={cs.CV}
}

@misc{kerm,
      title={{KERM}: Knowledge Enhanced Reasoning for Vision-and-Language Navigation}, 
      author={Xiangyang Li and Zihan Wang and Jiahao Yang and Yaowei Wang and Shuqiang Jiang},
      year={2023},
      eprint={2303.15796},
      archivePrefix={arXiv},
      primaryClass={cs.CV}
}

@inproceedings{qiao2022hop,
  title={{HOP}: History-and-order aware pre-training for vision-and-language navigation},
  author={Qiao, Yanyuan and Qi, Yuankai and Hong, Yicong and Yu, Zheng and Wang, Peng and Wu, Qi},
  booktitle={Proceedings of the IEEE/CVF Conference on Computer Vision and Pattern Recognition},
  pages={15418--15427},
  year={2022}
}

@inproceedings{hao2020towards,
  title={Towards learning a generic agent for vision-and-language navigation via pre-training},
  author={Hao, Weituo and Li, Chunyuan and Li, Xiujun and Carin, Lawrence and Gao, Jianfeng},
  booktitle={Proceedings of the IEEE/CVF conference on computer vision and pattern recognition},
  pages={13137--13146},
  year={2020}
}

@misc{liang2024cornav,
      title={{CorNav}: Autonomous Agent with Self-Corrected Planning for Zero-Shot Vision-and-Language Navigation}, 
      author={Xiwen Liang and Liang Ma and Shanshan Guo and Jianhua Han and Hang Xu and Shikui Ma and Xiaodan Liang},
      year={2024},
      eprint={2306.10322},
      archivePrefix={arXiv},
      primaryClass={cs.CV}
}

@inproceedings{schoenberger2016sfm,
    author={Sch\"{o}nberger, Johannes Lutz and Frahm, Jan-Michael},
    title={{Structure-from-Motion Revisited}},
    booktitle={Conference on Computer Vision and Pattern Recognition (CVPR)},
    year={2016}
}

@inproceedings{depthanything,
      title={{Depth Anything}: Unleashing the Power of Large-Scale Unlabeled Data}, 
      author={Yang, Lihe and Kang, Bingyi and Huang, Zilong and Xu, Xiaogang and Feng, Jiashi and Zhao, Hengshuang},
      booktitle={CVPR},
      year={2024}
}

@article{zhang2023recognize,
  title={{Recognize Anything}: A Strong Image Tagging Model},
  author={Zhang, Youcai and Huang, Xinyu and Ma, Jinyu and Li, Zhaoyang and Luo, Zhaochuan and Xie, Yanchun and Qin, Yuzhuo and Luo, Tong and Li, Yaqian and Liu, Shilong and others},
  journal={arXiv preprint arXiv:2306.03514},
  year={2023}
}

@misc{li2023blip2,
      title={{BLIP-2}: Bootstrapping Language-Image Pre-training with Frozen Image Encoders and Large Language Models}, 
      author={Junnan Li and Dongxu Li and Silvio Savarese and Steven Hoi},
      year={2023},
      eprint={2301.12597},
      archivePrefix={arXiv},
      primaryClass={cs.CV}
}

@misc{openai2022gpt4,
  title        = {{GPT-4}: Generative Pre-trained Transformer 4},
  author       = {OpenAI},
  year         = 2022,
  howpublished = {OpenAI API},
  url          = {https://beta.openai.com}
}

@inproceedings{scanqa,
  title={{ScanQA}: {3D} Question Answering for Spatial Scene Understanding},
  author={Azuma, Daichi and Miyanishi, Taiki and Kurita, Shuhei and Kawanabe, Motoaki},
  booktitle={Proceedings of the IEEE/CVF Conference on Computer Vision and Pattern Recognition (CVPR)},
  year={2022}
}

@misc{hong20233dllm,
      title={{3D-LLM}: Injecting the 3D World into Large Language Models}, 
      author={Yining Hong and Haoyu Zhen and Peihao Chen and Shuhong Zheng and Yilun Du and Zhenfang Chen and Chuang Gan},
      year={2023},
      eprint={2307.12981},
      archivePrefix={arXiv},
      primaryClass={cs.CV}
}

@article{liu2023grounding,
  title={{Grounding Dino}: Marrying dino with grounded pre-training for open-set object detection},
  author={Liu, Shilong and Zeng, Zhaoyang and Ren, Tianhe and Li, Feng and Zhang, Hao and Yang, Jie and Li, Chunyuan and Yang, Jianwei and Su, Hang and Zhu, Jun and others},
  journal={arXiv preprint arXiv:2303.05499},
  year={2023}
}

@inproceedings{li2023improving,
  title={Improving vision-and-language navigation by generating future-view image semantics},
  author={Li, Jialu and Bansal, Mohit},
  booktitle={Proceedings of the IEEE/CVF Conference on Computer Vision and Pattern Recognition},
  pages={10803--10812},
  year={2023}
}

@inproceedings{qiao2023vln,
  title={{VLN-PETL}: Parameter-Efficient Transfer Learning for Vision-and-Language Navigation},
  author={Qiao, Yanyuan and Yu, Zheng and Wu, Qi},
  booktitle={Proceedings of the IEEE/CVF International Conference on Computer Vision},
  pages={15443--15452},
  year={2023}
}

@inproceedings{an2023bevbert,
  title={{BEVbert}: Multimodal map pre-training for language-guided navigation},
  author={An, Dong and Qi, Yuankai and Li, Yangguang and Huang, Yan and Wang, Liang and Tan, Tieniu and Shao, Jing},
  booktitle={Proceedings of the IEEE/CVF International Conference on Computer Vision},
  pages={2737--2748},
  year={2023}
}

@inproceedings{wang2020environment,
  title={Environment-agnostic multitask learning for natural language grounded navigation},
  author={Wang, Xin Eric and Jain, Vihan and Ie, Eugene and Wang, William Yang and Kozareva, Zornitsa and Ravi, Sujith},
  booktitle={Computer Vision--ECCV 2020: 16th European Conference, Glasgow, UK, August 23--28, 2020, Proceedings, Part XXIV 16},
  pages={413--430},
  year={2020},
  organization={Springer}
}

@article{pan2023langnav,
  title={{LangNav}: Language as a perceptual representation for navigation},
  author={Pan, Bowen and Panda, Rameswar and Jin, SouYoung and Feris, Rogerio and Oliva, Aude and Isola, Phillip and Kim, Yoon},
  journal={arXiv preprint arXiv:2310.07889},
  year={2023}
}

@inproceedings{zhou2024navgpt,
  title={{NavGPT}: Explicit reasoning in vision-and-language navigation with large language models},
  author={Zhou, Gengze and Hong, Yicong and Wu, Qi},
  booktitle={Proceedings of the AAAI Conference on Artificial Intelligence},
  volume={38},
  number={7},
  pages={7641--7649},
  year={2024}
}

@article{lin2024navcot,
  title={{NavCoT}: Boosting LLM-Based Vision-and-Language Navigation via Learning Disentangled Reasoning},
  author={Lin, Bingqian and Nie, Yunshuang and Wei, Ziming and Chen, Jiaqi and Ma, Shikui and Han, Jianhua and Xu, Hang and Chang, Xiaojun and Liang, Xiaodan},
  journal={arXiv preprint arXiv:2403.07376},
  year={2024}
}

@article{jain2019stay,
  title={Stay on the path: Instruction fidelity in vision-and-language navigation},
  author={Jain, Vihan and Magalhaes, Gabriel and Ku, Alexander and Vaswani, Ashish and Ie, Eugene and Baldridge, Jason},
  journal={arXiv preprint arXiv:1905.12255},
  year={2019}
}

@article{ku2020room,
  title={Room-across-room: Multilingual vision-and-language navigation with dense spatiotemporal grounding},
  author={Ku, Alexander and Anderson, Peter and Patel, Roma and Ie, Eugene and Baldridge, Jason},
  journal={arXiv preprint arXiv:2010.07954},
  year={2020}
}

@inproceedings{xia2018gibson,
  title={{Gibson Env}: Real-world perception for embodied agents},
  author={Xia, Fei and Zamir, Amir R and He, Zhiyang and Sax, Alexander and Malik, Jitendra and Savarese, Silvio},
  booktitle={Proceedings of the IEEE conference on computer vision and pattern recognition},
  pages={9068--9079},
  year={2018}
}

@article{ramakrishnan2021habitat,
  title={{Habitat-Matterport 3D dataset (HM3D)}: 1000 large-scale 3d environments for embodied ai},
  author={Ramakrishnan, Santhosh K and Gokaslan, Aaron and Wijmans, Erik and Maksymets, Oleksandr and Clegg, Alex and Turner, John and Undersander, Eric and Galuba, Wojciech and Westbury, Andrew and Chang, Angel X and others},
  journal={arXiv preprint arXiv:2109.08238},
  year={2021}
}

@inproceedings{schoenberger2016mvs,
    author={Sch\"{o}nberger, Johannes Lutz and Zheng, Enliang and Pollefeys, Marc and Frahm, Jan-Michael},
    title={Pixelwise View Selection for Unstructured Multi-View Stereo},
    booktitle={European Conference on Computer Vision (ECCV)},
    year={2016},
}

@article{zhang2024navid,
  title={{NaVid}: Video-based VLM Plans the Next Step for Vision-and-Language Navigation},
  author={Zhang, Jiazhao and Wang, Kunyu and Xu, Rongtao and Zhou, Gengze and Hong, Yicong and Fang, Xiaomeng and Wu, Qi and Zhang, Zhizheng and He, Wang},
  journal={arXiv preprint arXiv:2402.15852},
  year={2024}
}

@inproceedings{hwang2023meta,
  title={{Meta-explore}: Exploratory hierarchical vision-and-language navigation using scene object spectrum grounding},
  author={Hwang, Minyoung and Jeong, Jaeyeon and Kim, Minsoo and Oh, Yoonseon and Oh, Songhwai},
  booktitle={Proceedings of the IEEE/CVF Conference on Computer Vision and Pattern Recognition},
  pages={6683--6693},
  year={2023}
}

@inproceedings{gao2023adaptive,
  title={Adaptive zone-aware hierarchical planner for vision-language navigation},
  author={Gao, Chen and Peng, Xingyu and Yan, Mi and Wang, He and Yang, Lirong and Ren, Haibing and Li, Hongsheng and Liu, Si},
  booktitle={Proceedings of the IEEE/CVF Conference on Computer Vision and Pattern Recognition},
  pages={14911--14920},
  year={2023}
}

@article{Matterport3D,
  title={Matterport3D: Learning from RGB-D Data in Indoor Environments},
  author={Chang, Angel and Dai, Angela and Funkhouser, Thomas and Halber, Maciej and Niessner, Matthias and Savva, Manolis and Song, Shuran and Zeng, Andy and Zhang, Yinda},
  journal={International Conference on 3D Vision (3DV)},
  year={2017}
}

@misc{wikiwalking,
  title        = {Preferred walking speed},
  author       = {Wikipedia},
  howpublished = {\url{https://en.wikipedia.org/wiki/Preferred_walking_speed}}
}

@article{tan2019lxmert,
  title={Lxmert: Learning cross-modality encoder representations from transformers},
  author={Tan, Hao and Bansal, Mohit},
  journal={arXiv preprint arXiv:1908.07490},
  year={2019}
}

@misc{vicuna2023,
    title = {Vicuna: An Open-Source Chatbot Impressing GPT-4 with 90\%* ChatGPT Quality},
    url = {https://lmsys.org/blog/2023-03-30-vicuna/},
    author = {Chiang, Wei-Lin and Li, Zhuohan and Lin, Zi and Sheng, Ying and Wu, Zhanghao and Zhang, Hao and Zheng, Lianmin and Zhuang, Siyuan and Zhuang, Yonghao and Gonzalez, Joseph E. and Stoica, Ion and Xing, Eric P.},
    month = {March},
    year = {2023}
}

@article{chen2024mapgpt,
  title={MapGPT: Map-Guided Prompting for Unified Vision-and-Language Navigation},
  author={Chen, Jiaqi and Lin, Bingqian and Xu, Ran and Chai, Zhenhua and Liang, Xiaodan and Wong, Kwan-Yee K},
  journal={arXiv preprint arXiv:2401.07314},
  year={2024}
}

@misc{liu2023llava,
      title={Visual Instruction Tuning}, 
      author={Liu, Haotian and Li, Chunyuan and Wu, Qingyang and Lee, Yong Jae},
      publisher={NeurIPS},
      year={2023},
}

@article{long2023discuss,
  title={Discuss before moving: Visual language navigation via multi-expert discussions},
  author={Long, Yuxing and Li, Xiaoqi and Cai, Wenzhe and Dong, Hao},
  journal={arXiv preprint arXiv:2309.11382},
  year={2023}
}

@inproceedings{dbscan,
author = {Ester, Martin and Kriegel, Hans-Peter and Sander, J\"{o}rg and Xu, Xiaowei},
title = {A density-based algorithm for discovering clusters in large spatial databases with noise},
year = {1996},
publisher = {AAAI Press},
booktitle = {Proceedings of the Second International Conference on Knowledge Discovery and Data Mining},
pages = {226–231},
numpages = {6},
location = {Portland, Oregon},
series = {KDD'96}
}

@inproceedings{liu2024volumetric,
  title={Volumetric Environment Representation for Vision-Language Navigation},
  author={Liu, Rui and Wang, Wenguan and Yang, Yi},
  booktitle={CVPR},
  pages={16317--16328},
  year={2024}
}

@InProceedings{Wang2024GOAT,
    author    = {Wang, Liuyi and He, Zongtao and Dang, Ronghao and Shen, Mengjiao and Liu, Chengju and Chen, Qijun},
    title     = {Vision-and-Language Navigation via Causal Learning},
    booktitle = {The IEEE/CVF Conference on Computer Vision and Pattern Recognition (CVPR)},
    year      = {2024}
}

@misc{zhao2024overnavelevatingiterativevisionandlanguage,
      title={OVER-NAV: Elevating Iterative Vision-and-Language Navigation with Open-Vocabulary Detection and StructurEd Representation}, 
      author={Ganlong Zhao and Guanbin Li and Weikai Chen and Yizhou Yu},
      year={2024},
      eprint={2403.17334},
      archivePrefix={arXiv},
      primaryClass={cs.CV},
      url={https://arxiv.org/abs/2403.17334}, 
}

@article{li2023panogen,
  title={Panogen: Text-conditioned panoramic environment generation for vision-and-language navigation},
  author={Li, Jialu and Bansal, Mohit},
  journal={Advances in Neural Information Processing Systems},
  volume={36},
  pages={21878--21894},
  year={2023}
}

@INPROCEEDINGS{depth_search,
  author={Tarjan, Robert},
  booktitle={12th Annual Symposium on Switching and Automata Theory (swat 1971)}, 
  title={Depth-first search and linear graph algorithms}, 
  year={1971},
  volume={},
  number={},
  pages={114-121},
  doi={10.1109/SWAT.1971.10}}

@inproceedings{zhou2025navgpt,
  title={Navgpt-2: Unleashing navigational reasoning capability for large vision-language models},
  author={Zhou, Gengze and Hong, Yicong and Wang, Zun and Wang, Xin Eric and Wu, Qi},
  booktitle={European Conference on Computer Vision},
  pages={260--278},
  year={2025},
  organization={Springer}
}

@inproceedings{han2025roomtour3d,
  title={Roomtour3d: Geometry-aware video-instruction tuning for embodied navigation},
  author={Han, Mingfei and Ma, Liang and Zhumakhanova, Kamila and Radionova, Ekaterina and Zhang, Jingyi and Chang, Xiaojun and Liang, Xiaodan and Laptev, Ivan},
  booktitle={Proceedings of the Computer Vision and Pattern Recognition Conference},
  pages={27586--27596},
  year={2025}
}

@article{hendrycks2019benchmarking,
  title={Benchmarking neural network robustness to common corruptions and perturbations},
  author={Hendrycks, Dan and Dietterich, Thomas},
  journal={arXiv preprint arXiv:1903.12261},
  year={2019}
}

@article{hao2025conav,
  title={CoNav: Collaborative Cross-Modal Reasoning for Embodied Navigation},
  author={Hao, Haihong and Han, Mingfei and Li, Changlin and Li, Zhihui and Chang, Xiaojun},
  journal={arXiv preprint arXiv:2505.16663},
  year={2025}
}

@article{wang2024bootstrapping,
  title={Bootstrapping Language-Guided Navigation Learning with Self-Refining Data Flywheel},
  author={Wang, Zun and Li, Jialu and Hong, Yicong and Li, Songze and Li, Kunchang and Yu, Shoubin and Wang, Yi and Qiao, Yu and Wang, Yali and Bansal, Mohit and others},
  journal={arXiv preprint arXiv:2412.08467},
  year={2024}
}

@inproceedings{wang2024vision,
  title={Vision-and-language navigation via causal learning},
  author={Wang, Liuyi and He, Zongtao and Dang, Ronghao and Shen, Mengjiao and Liu, Chengju and Chen, Qijun},
  booktitle={Proceedings of the IEEE/CVF Conference on Computer Vision and Pattern Recognition},
  pages={13139--13150},
  year={2024}
}

@article{zhang2024agent,
  title={Agent Journey Beyond RGB: Unveiling Hybrid Semantic-Spatial Environmental Representations for Vision-and-Language Navigation},
  author={Zhang, Xuesong and Xu, Yunbo and Li, Jia and Hu, Zhenzhen and Hong, Richnag},
  journal={arXiv preprint arXiv:2412.06465},
  year={2024}
}

@inproceedings{liu2023bird,
  title={Bird's-Eye-View Scene Graph for Vision-Language Navigation},
  author={Liu, Rui and Wang, Xiaohan and Wang, Wenguan and Yang, Yi},
  booktitle={Proceedings of the IEEE/CVF International Conference on Computer Vision},
  pages={10968--10980},
  year={2023}
}

@inproceedings{li2022bevformer,
  title={Bevformer: Learning bird’s-eye-view representation from multi-camera images via spatiotemporal transformers},
  author={Li, Zhiqi and Wang, Wenhai and Li, Hongyang and Xie, Enze and Sima, Chonghao and Lu, Tong and Qiao, Yu and Dai, Jifeng},
  booktitle={European conference on computer vision},
  pages={1--18},
  year={2022},
  organization={Springer}
}

@article{zhou2024same,
  title={SAME: Learning Generic Language-Guided Visual Navigation with State-Adaptive Mixture of Experts},
  author={Zhou, Gengze and Hong, Yicong and Wang, Zun and Zhao, Chongyang and Bansal, Mohit and Wu, Qi},
  journal={arXiv preprint arXiv:2412.05552},
  year={2024}
}

@inproceedings{wang2025continuous,
  title={Continuous 3d perception model with persistent state},
  author={Wang, Qianqian and Zhang, Yifei and Holynski, Aleksander and Efros, Alexei A and Kanazawa, Angjoo},
  booktitle={Proceedings of the Computer Vision and Pattern Recognition Conference},
  pages={10510--10522},
  year={2025}
}

@article{fan2025vlm,
  title={VLM-3R: Vision-Language Models Augmented with Instruction-Aligned 3D Reconstruction},
  author={Fan, Zhiwen and Zhang, Jian and Li, Renjie and Zhang, Junge and Chen, Runjin and Hu, Hezhen and Wang, Kevin and Qu, Huaizhi and Wang, Dilin and Yan, Zhicheng and others},
  journal={arXiv preprint arXiv:2505.20279},
  year={2025}
}

@inproceedings{wang2025vggt,
  title={Vggt: Visual geometry grounded transformer},
  author={Wang, Jianyuan and Chen, Minghao and Karaev, Nikita and Vedaldi, Andrea and Rupprecht, Christian and Novotny, David},
  booktitle={Proceedings of the Computer Vision and Pattern Recognition Conference},
  pages={5294--5306},
  year={2025}
}

@inproceedings{agarwal2010bundle,
  title={Bundle adjustment in the large},
  author={Agarwal, Sameer and Snavely, Noah and Seitz, Steven M and Szeliski, Richard},
  booktitle={Computer Vision--ECCV 2010: 11th European Conference on Computer Vision, Heraklion, Crete, Greece, September 5--11, 2010, Proceedings, Part II},
  pages={29--42},
  year={2010},
  organization={Springer}
}

@article{agarwal2011rome,
  title={Building Rome in a day},
  author={Agarwal, Sameer and Furukawa, Yasutaka and Snavely, Noah and Simon, Ian and Curless, Brian and Seitz, Steven M and Szeliski, Richard},
  journal={Communications of the ACM},
  volume={54},
  number={10},
  pages={105--112},
  year={2011},
  publisher={ACM}
}

@book{hartley2003multiple,
  title={Multiple View Geometry in Computer Vision},
  author={Hartley, Richard and Zisserman, Andrew},
  year={2003},
  publisher={Cambridge University Press}
}

@article{lowe2004distinctive,
  title={Distinctive image features from scale-invariant keypoints},
  author={Lowe, David G},
  journal={International Journal of Computer Vision},
  volume={60},
  pages={91--110},
  year={2004},
  publisher={Springer}
}

@inproceedings{schonberger2016sfm,
  title={Structure-from-Motion revisited},
  author={Sch{\"o}nberger, Johannes Lutz and Frahm, Jan-Michael},
  booktitle={Proceedings of the IEEE Conference on Computer Vision and Pattern Recognition (CVPR)},
  pages={4104--4113},
  year={2016},
  organization={IEEE}
}

@inproceedings{seitz2006comparison,
  title={A comparison and evaluation of multi-view stereo reconstruction algorithms},
  author={Seitz, Steven M and Curless, Brian and Diebel, James and Scharstein, Daniel and Szeliski, Richard},
  booktitle={2006 IEEE Computer Society Conference on Computer Vision and Pattern Recognition (CVPR'06)},
  volume={1},
  pages={519--528},
  year={2006},
  organization={IEEE}
}

@inproceedings{snavely2006photo,
  title={Photo tourism: exploring photo collections in 3D},
  author={Snavely, Noah and Seitz, Steven M and Szeliski, Richard},
  booktitle={ACM SIGGRAPH 2006 Papers},
  pages={835--846},
  year={2006},
  organization={ACM}
}

@article{snavely2008modeling,
  title={Modeling the world from Internet photo collections},
  author={Snavely, Noah and Seitz, Steven M and Szeliski, Richard},
  journal={International Journal of Computer Vision},
  volume={80},
  pages={189--210},
  year={2008},
  publisher={Springer}
}

@inproceedings{triggs2000bundle,
  title={Bundle adjustment---a modern synthesis},
  author={Triggs, Bill and McLauchlan, Philip F and Hartley, Richard I and Fitzgibbon, Andrew W},
  booktitle={Vision Algorithms: Theory and Practice: International Workshop on Vision Algorithms, Corfu, Greece, September 21--22, 1999, Proceedings},
  pages={298--372},
  year={2000},
  organization={Springer}
}

@article{cadena2016past,
  title={Past, present, and future of simultaneous localization and mapping: Toward the robust-perception age},
  author={Cadena, Cesar and Carlone, Luca and Carrillo, Henry and Latif, Yasir and Scaramuzza, Davide and Neira, Jose and Reid, Ian and Leonard, John J},
  journal={IEEE Transactions on Robotics},
  volume={32},
  number={6},
  pages={1309--1332},
  year={2016},
  publisher={IEEE}
}

@article{davison2007monoslam,
  title={Monoslam: Real-time single camera slam},
  author={Davison, Andrew J and Reid, Ian D and Molton, Nicholas D and Stasse, Olivier},
  journal={IEEE Transactions on Pattern Analysis and Machine Intelligence},
  volume={29},
  number={6},
  pages={1052--1067},
  year={2007},
  publisher={IEEE}
}

@article{durrant2006slam1,
  title={Simultaneous localization and mapping: Part I},
  author={Durrant-Whyte, Hugh and Bailey, Tim},
  journal={IEEE Robotics \& Automation Magazine},
  volume={13},
  number={2},
  pages={99--110},
  year={2006},
  publisher={IEEE}
}

@inproceedings{engel2014lsdslam,
  title={LSD-SLAM: Large-scale direct monocular SLAM},
  author={Engel, Jakob and Sch{\"o}ps, Thomas and Cremers, Daniel},
  booktitle={European Conference on Computer Vision},
  pages={834--849},
  year={2014},
  organization={Springer}
}

@inproceedings{klein2007ptam,
  title={Parallel tracking and mapping for small AR workspaces},
  author={Klein, Georg and Murray, David},
  booktitle={2007 6th IEEE and ACM International Symposium on Mixed and Augmented Reality},
  pages={225--234},
  year={2007},
  organization={IEEE}
}

@article{mur2015orbslam,
  title={ORB-SLAM: A versatile and accurate monocular SLAM system},
  author={Mur-Artal, Raul and Montiel, Jose Maria Martinez and Tard{\'o}s, Juan D},
  journal={IEEE Transactions on Robotics},
  volume={31},
  number={5},
  pages={1147--1163},
  year={2015},
  publisher={IEEE}
}

@inproceedings{newcombe2011dtam,
  title={DTAM: Dense tracking and mapping in real-time},
  author={Newcombe, Richard A and Lovegrove, Steven J and Davison, Andrew J},
  booktitle={2011 International Conference on Computer Vision},
  pages={2320--2327},
  year={2011},
  organization={IEEE}
}

@inproceedings{chen2022tensorf,
  title={TensoRF: Tensorial radiance fields},
  author={Chen, Anpei and Xu, Zexiang and Geiger, Andreas and Yu, Jingyi and Su, Hao},
  booktitle={European Conference on Computer Vision},
  pages={333--350},
  year={2022},
  organization={Springer}
}

@inproceedings{fridovich2022plenoxels,
  title={Plenoxels: Radiance fields without neural networks},
  author={Fridovich-Keil, Sara and Yu, Alex and Tancik, Matthew and Chen, Qinhong and Recht, Benjamin and Kanazawa, Angjoo},
  booktitle={Proceedings of the IEEE/CVF Conference on Computer Vision and Pattern Recognition},
  pages={5501--5510},
  year={2022},
  organization={IEEE}
}

@article{mildenhall2021nerf,
  title={NeRF: Representing scenes as neural radiance fields for view synthesis},
  author={Mildenhall, Ben and Srinivasan, Pratul P and Tancik, Matthew and Barron, Jonathan T and Ramamoorthi, Ravi and Ng, Ren},
  journal={Communications of the ACM},
  volume={65},
  number={1},
  pages={99--106},
  year={2021},
  publisher={ACM}
}

@article{mueller2022instant,
  title={Instant neural graphics primitives with a multiresolution hash encoding},
  author={M{\"u}ller, Thomas and Evans, Alex and Schied, Christoph and Keller, Alexander},
  journal={ACM Transactions on Graphics (TOG)},
  volume={41},
  number={4},
  pages={1--15},
  year={2022},
  publisher={ACM}
}

@article{wang2021neus,
  title={NeuS: Learning neural implicit surfaces by volume rendering for multi-view reconstruction},
  author={Wang, Peng and Liu, Lingjie and Liu, Yuan and Theobalt, Christian and Komura, Taku and Wang, Wenping},
  journal={arXiv preprint arXiv:2106.10689},
  year={2021}
}

@article{kerbl2023gaussiansplatting,
  title={3D Gaussian Splatting for Real-Time Radiance Field Rendering},
  author={Kerbl, Bernhard and Kopanas, Georgios and Leimk{\"u}hler, Thomas and Drettakis, George},
  journal={ACM Transactions on Graphics (TOG)},
  volume={42},
  number={4},
  pages={139:1--139:1},
  year={2023},
  publisher={ACM}
}

@inproceedings{wang2024dust3r,
  title={DUSt3R: Geometric 3D vision made easy},
  author={Wang, Shuzhe and Leroy, Vincent and Cabon, Yohann and Chidlovskii, Boris and Revaud, Jerome},
  booktitle={Proceedings of the IEEE/CVF Conference on Computer Vision and Pattern Recognition (CVPR)},
  pages={20697--20709},
  year={2024},
  organization={IEEE}
}

@inproceedings{wang2025moge,
  title={Moge: Unlocking accurate monocular geometry estimation for open-domain images with optimal training supervision},
  author={Wang, Ruicheng and Xu, Sicheng and Dai, Cassie and Xiang, Jianfeng and Deng, Yu and Tong, Xin and Yang, Jiaolong},
  booktitle={Proceedings of the Computer Vision and Pattern Recognition Conference},
  pages={5261--5271},
  year={2025}
}

@inproceedings{long2024discuss,
  title={Discuss before moving: Visual language navigation via multi-expert discussions},
  author={Long, Yuxing and Li, Xiaoqi and Cai, Wenzhe and Dong, Hao},
  booktitle={2024 IEEE International Conference on Robotics and Automation (ICRA)},
  pages={17380--17387},
  year={2024},
  organization={IEEE}
}

@article{Tosi2021SMDNetsSM,
  title={SMD-Nets: Stereo Mixture Density Networks},
  author={Fabio Tosi and Yiyi Liao and Carolin Schmitt and Andreas Geiger},
  journal={2021 IEEE/CVF Conference on Computer Vision and Pattern Recognition (CVPR)},
  year={2021},
  pages={8938-8948},
  url={https://api.semanticscholar.org/CorpusID:233182047}
}
\begin{IEEEbiography}
[{\includegraphics[width=1in,height=1.25in,clip,keepaspectratio]{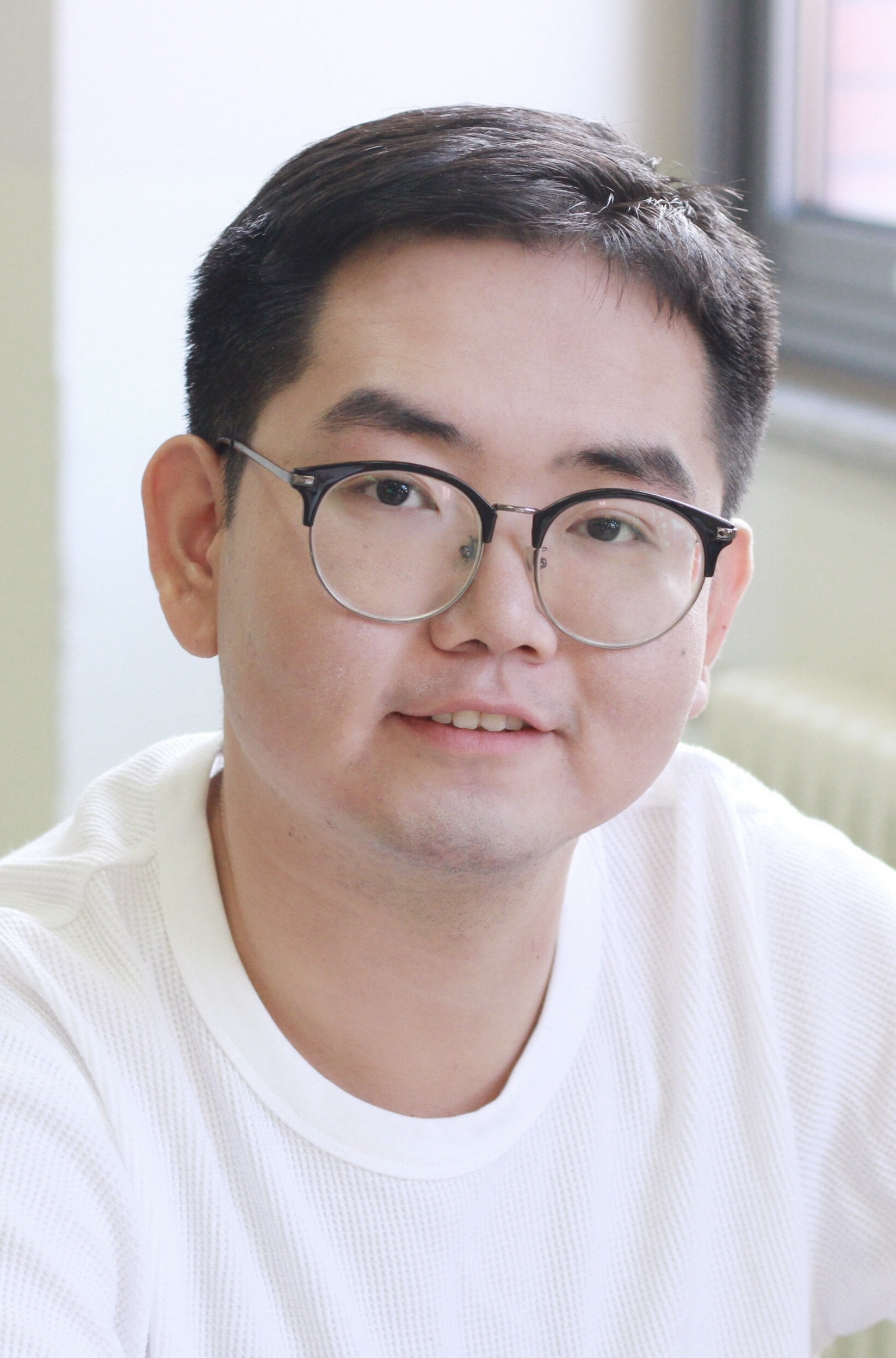}}]{Mingfei Han} is currently a postdoctoral associate at Mohamed Bin Zayed University of Artificial Intelligence. He obtained his Ph.D. degree from University of Technology Sydney. He received the B.Eng. degree from Nankai University and the M.Eng. degree from University of Chinese Academy of Sciences. His research interests lie in computer vision and machine learning, with a particular emphasis on large vision-language models, video object perception and their applications in robotics.
\end{IEEEbiography}

\vspace{-.8 em}

\begin{IEEEbiography}
[{\includegraphics[width=1in,height=1.25in,clip,keepaspectratio]{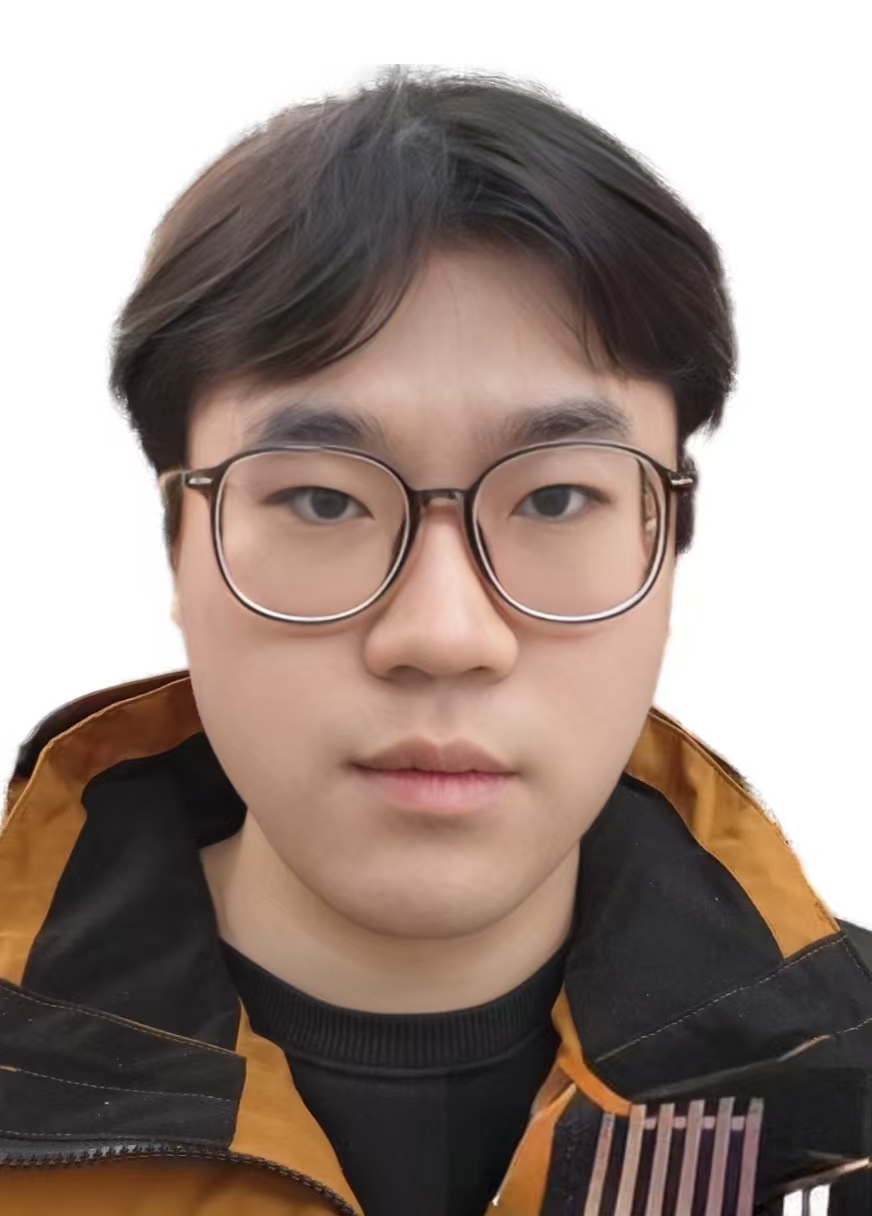}}]{Haihong Hao} is currently a Ph.D. student at the School of Information Science and Technology (USTC). He received his B.E. degree in the School of Computer and Artificial Intelligence, Zhengzhou University, China.
His research
interests include computer vision and multimedia
content analysis.
\end{IEEEbiography}

\vspace{-.8 em}

\begin{IEEEbiography}
[{\includegraphics[width=1in,height=1.25in,clip,keepaspectratio]{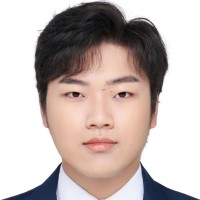}}]{Liang Ma} is currently a master student at Department of Computer Vision, Mohamed bin Zayed University of Artificial Intelligence (MBZUAI). He received his B.E. degree in the School of Computer and Artificial Intelligence, Sun Yat-sen University, China.
His research
interests include computer vision and multimedia
content analysis.
\end{IEEEbiography}

\vspace{-.8 em}

\begin{IEEEbiography}
[{\includegraphics[width=1in,height=1.25in,clip,keepaspectratio]{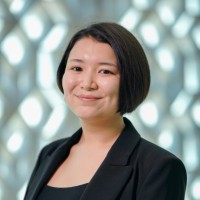}}]{Kamila Zhumakhanova} is a Master’s student in the Department of Computer Vision at the Mohamed bin Zayed University of Artificial Intelligence (MBZUAI). Her research interests include computer vision and machine learning, with a focus on representation learning and visual understanding.
\end{IEEEbiography}
\vspace{-.8 em}

\begin{IEEEbiography}
[{\includegraphics[width=1in,height=1.25in,clip,keepaspectratio]{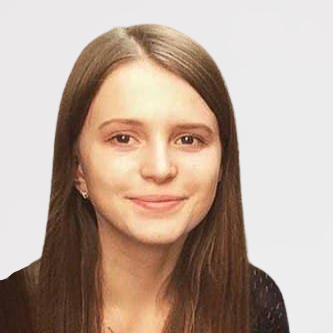}}]{Ekaterina Radionova} is a Research Scientist in the Department of Computer Vision at the Mohamed bin Zayed University of Artificial Intelligence (MBZUAI). Her research interests include computer vision and machine learning, with a focus on representation learning and visual understanding.
\end{IEEEbiography}
\vspace{-.8 em}

\begin{IEEEbiographynophoto}{Jingyi Zhang} is a master student with Shenzhen Campus of Sun Yat-Sen University. His research interests include computer vision and machine learning, with a focus on representation learning and visual understanding.
\end{IEEEbiographynophoto}
\vspace{-.8 em}

\begin{IEEEbiography}[{\includegraphics[width=1in,height=1.25in,clip,keepaspectratio]{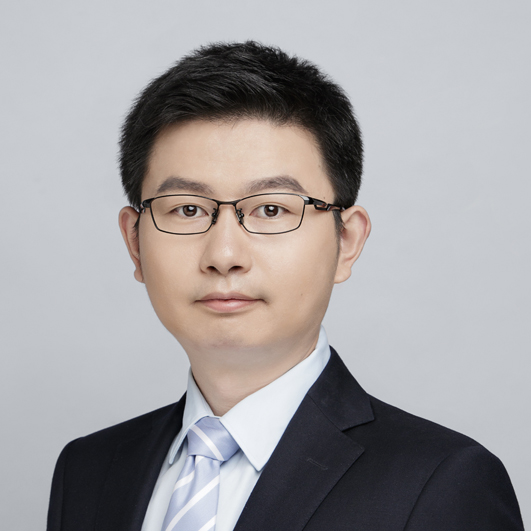}}]{Xiaojun Chang} (Senior Member, IEEE) is a Professor at Future Media Computing Lab, School of Information Science
and Technology, University of Science and Technology of China. He is also a visiting Professor with Department of Computer Vision, Mohamed bin Zayed University of Artificial Intelligence (MBZUAI). He
has spent most of his time working on exploring multiple signals (visual,
acoustic, textual) for automatic content analysis in unconstrained or
surveillance videos. He has achieved top performances in various
international competitions, such as TRECVID MED, TRECVID SIN, and
TRECVID AVS. \end{IEEEbiography}
\vspace{-.8 em}

\begin{IEEEbiography}
[{\includegraphics[width=1in,height=1.25in,clip,keepaspectratio]{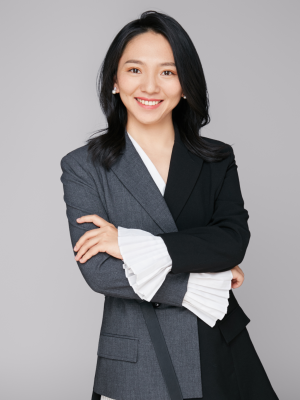}}]{Xiaodan Liang} is an Associate Professor in the Department of Computer Vision at the Mohamed bin Zayed University of Artificial Intelligence (MBZUAI), and a joint Professor at Sun Yat-sen Univerisity, China. She was a Project Scientist at Carnegie Mellon University, working with Prof. Eric Xing. She has published over 120 cutting-edge papers on visual-language understanding and generation, and its application on embodied AI, which have appeared in the most prestigious journals and conferences in the field, with Google Citation 30000+. She serves as regular Area Chairs of ICCV, CVPR, NeurIPS, ICML, ICLR and AAAI regularly, and Tutorial Chair of CVPR 2021, Ombud chair of CVPR 2023, Local chairs of ICCV 2029. She has been awarded ACM China and CCF Best Doctoral Dissertation Award and Alibaba DAMO Academy Young Fellow. Her research has been applied in the key products in several renowned AI companies such as Deepseek, Lenovo, ByteDance and Tencent Inc.
\end{IEEEbiography}
\vspace{-.8 em}

\begin{IEEEbiography}
[{\includegraphics[width=1in,height=1.25in,clip,keepaspectratio]{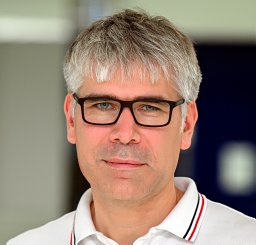}}]{Ivan Laptev} received the PhD degree in computer science from the Royal Institute of Technology, in 2004 and the habilitation degree from École Normale Supérieure, in 2013. He is a visiting professor with MBZUAI and a senior researcher on leave from Inria. His main research interests include visual recognition of human actions, objects and interactions, and more recently robotics. He has published more than 150 technical papers most of which appeared in international journals and major peer-reviewed conferences of the field. He served/serves as a program chair for CVPR’18, ICCV’23, and ACCV’24. He received an ERC Starting grant and was awarded a Helmholtz prize for significant impact on computer vision.
\end{IEEEbiography}
\vspace{-.8 em}

\end{document}